\documentclass[11pt,a4paper]{article}

\pdfoutput=1

\usepackage{amsmath,amsthm,verbatim,amssymb,amsfonts,amscd,graphicx}
\usepackage{graphics}
\usepackage{bm}
\usepackage{enumerate}
\usepackage{lscape}
\usepackage{hyperref}
\usepackage{cleveref}
\usepackage{algorithm}
\usepackage{algorithmicx}
\usepackage{algpseudocode}



\DeclareMathOperator{\tr}{tr}
\DeclareMathOperator{\ve}{vec}
\DeclareMathOperator{\cov}{cov}

\theoremstyle{plain}

\theoremstyle{definition}
\newtheorem{definition}{Definition}

\DeclareMathOperator*{\argmin}{arg\,min}

\algnewcommand{\Inputs}[1]{%
  \State \textbf{Inputs:}
  \Statex \hspace*{\algorithmicindent}\parbox[t]{.8\linewidth}{\raggedright #1}
}
\algnewcommand{\Initialize}[1]{%
  \State \textbf{Initialize:}
  \Statex \hspace*{\algorithmicindent}\parbox[t]{.8\linewidth}{\raggedright #1}
}
\algnewcommand{\Outputs}[1]{%
  \State \textbf{Outputs:}
  \Statex \hspace*{\algorithmicindent}\parbox[t]{.8\linewidth}{\raggedright #1}
}

\begin{document}

\title{Information Mandala: Statistical Distance Matrix with Clustering}

\author{Xin Lu \footnotemark[1]}

\date{}

\maketitle

\begin{abstract}
In machine learning, observation features are measured in a metric space to obtain their distance function for optimization. Given similar features that are statistically sufficient as a population, a statistical distance between two probability distributions can be calculated for more precise learning. Provided the observed features are multi-valued, the statistical distance function is still efficient. However, due to its scalar output, it cannot be applied to represent detailed distances between feature elements. To resolve this problem, this paper extends the traditional statistical distance to a matrix form, called a statistical distance matrix. In experiments, the proposed approach performs well in object recognition tasks and clearly and intuitively represents the dissimilarities between cat and dog images in the CIFAR dataset, even when directly calculated using the image pixels. By using the hierarchical clustering of the statistical distance matrix, the image pixels can be separated into several clusters that are geometrically arranged around a center like a Mandala pattern. The statistical distance matrix with clustering, called the Information Mandala, is beyond ordinary saliency maps and can help to understand the basic principles of the convolution neural network.
\end{abstract}

\footnotetext[1]{Iwate University, Japan.}

\section{Introduction}\label{sec:introduction}

Classification is a type of supervised learning in machine learning that identifies to which of a set of categories a new observation belongs, based on a training set of labeled observations. The corresponding procedure for unsupervised learning is called clustering, which groups observations into categories based on their inherent similarities. In both classification and clustering, observation features are measured in a metric space, and their dissimilarities or distances are calculated for optimization. For example, a support vector machine (SVM) \cite{CortesV95} in classification needs to measure the distance between two observation categories using the most efficient kernel function. For clustering, the K-means \cite{MacQueen67}  approach aims to divide observations into categories to minimize the within-cluster sum of squares metric of the features in Euclidean or Mahalanobis space.

Provided the observation features are considered random variables or the feature set is considered a random vector in a probability space, measuring the distance between observations can be interpreted as quantifying a statistical distance between two probability distributions. Statistical distances have special mathematical properties that not all distances have. These properties include making distance measurements not only more effective and appropriate but also more robust to small outliers. Some important statistical distances, such as the Mahalanobis distance \cite{Mahalanobis36}, Bhattacharyya distance \cite{Bhattacharyya43}, Hellinger distance \cite{Hellinger09}, Kullback-Leibler divergence \cite{Kullback51,Kullback59}, and Chernoff distance \cite{Chernoff52} , have been applied to artificial intelligence applications, such as image segmentation \cite{Ayed10, Kang13}, texture segmentation \cite{Reyes06}, color and texture matching \cite{Derpanis12-1}, feature extraction \cite{Shirosawa17}, speech recognition \cite{You09}, and action recognition \cite{Derpanis12-2}. However, a clear limitation of these general statistical distances is that they only provide a scalar output to represent a global feature distance between two observations, regardless of the size of the feature set or the dimensionality of the corresponding random vector. Thus, the local distances of all features in the set or the relationships of all elements in the random vector cannot be elaborated. Therefore, an important problem is how to refine the concept of statistical distance to move from a scalar to a matrix.

The concept of a distance matrix has been introduced in graph theory \cite{Harary65}. In a directed graph, a distance matrix is defined by a weighted adjacency matrix. Given that each edge is assigned a weight, the distance between two vertices can be defined as the minimum sum of the weights of the shortest paths connecting the two vertices. The distance matrix is asymmetric and not metric because the paths are oriented. If there are enough samples of each vertex, the correlation matrix or partial correlation matrix is used to identify the weights and quantify the distance matrix. However, due to its information loss from the assumption that all the data is in a probability space, the correlation matrix is not yet delicate or precise enough to satisfy some machine learning requirements.

To solve the above problems, this paper transforms traditional statistical distances into their matrix forms through a simple de-trace operation, and experimentally demonstrates the results for complicated distance performances using the CIFAR-10 dataset, which is the most famous dataset in machine learning.

\section{Preliminaries}\label{sec:preliminary}

First, we provide some definitions of probability theory to specify the statistical distance.

\begin{definition}\label{definition1}
Let $(\Omega, \mathcal{F}, P)$ be a probability space, where $\Omega := \{\omega\}$ is a set of sample points. $\mathcal{F}$, which is a Borel $\sigma$-field (or $\sigma$-algebra), is called a collection of subsets of a sample space $\Omega$. An element $A \in \mathcal{F}$ is referred to as a measurable set in measure theory, and an event in probability and statistics. $P(A)$ are probability measures on the measurable space $(\Omega, \mathcal{F})$.
\end{definition}

\begin{definition}\label{definition3}
In probability theory, a measurable function from a probability space $(\Omega, \mathcal{F}, P)$ to a measurable space $(\Lambda, \mathcal{G})$ is called a $(\Lambda, \mathcal{G})$-valued random variable, and is denoted by one of $X$, $Y$, $Z$, $\ldots$. Let $\mathcal{G} := \mathcal{B}(\Lambda)$ (Borel $\sigma$-field). If $X$ is a measurable function from $(\Omega, \mathcal{F}, P)$ to $(\mathbb{R}, \mathcal{B}(\mathbb{R}))$, it is called a real-valued random variable. If $X := [X_{1}, \ldots, X_{d}]^\intercal$ is a measurable function from $(\Omega, \mathcal{F}, P)$ to $(\mathbb{R}^{d}, \mathcal{B}(\mathbb{R}^{d}))$, it is called a $d$-dimensional random vector, where $X_{i}$ is the $i$-th component of $X$, and $X_{1},\ \ldots\ ,X_{d}$ are random variables on a common probability space.
\end{definition}

\begin{definition}\label{definition4}
Let $X$ be a $(\Lambda, \mathcal{G})$-valued random variable on a probability space $(\Omega, \mathcal{F}, P)$. Then, a probability measure $P \circ X^{-1}$ on a measurable space $(\Lambda, \mathcal{G})$  is defined as $P \circ X^{-1}(B) := P(X^{-1}(B)) = P(X \in B), B \in \mathcal{G}$. Then $P \circ X^{-1}$ is called the distribution of $X$, and is denoted by $P_{X}$.
\end{definition}

\begin{definition}\label{definition7}
Consider a real-valued random variable or random vector $X$ with distribution $P_{X}$ on a measurable space $(\mathbb{R},\mathcal{B}(\mathbb{R}))$. If $F(x) := P(X \leq x) = P_{X}((-\infty,x])$, for $x \in \mathbb{R}$, then $F(x)$ is called the cumulative distribution function of $X$ on $\mathbb{R}$. Moreover, if $F(x)$ is absolutely continuous on $\mathbb{R}$, then $f_{X}(x) := dF(x)/dx$ is called the probability density function of $X$.
\end{definition}

\begin{definition}\label{definition8}
Generally, for a $d$-dimensional random vector $X_1$ with distribution $P_{X}$ on the measurable space $(\mathbb{R}^{d},\mathcal{B}(\mathbb{R}^{d}))$, the cumulative distribution function is defined as $F_1(\bm{x}) := P(X_1 \leq \bm{x}) = P_{X}((-\infty,\bm{x}])$, for $\bm{x} \in \mathbb{R}^{d}$. If $F_1(\bm{x})$ is absolutely continuous with respect to $\bm{x}$ on $\mathbb{R}^{d}$, then $p_{X}(\bm{x}) := dF_1(\bm{x})/d\bm{x}$ is called the probability density function of $X_1$. Given another probability space $(\Omega, \mathcal{F}, Q)$, for a $d$-dimensional random vector $X_2$ with distribution $Q_{X}$ on the measurable space $(\mathbb{R}^{d},\mathcal{B}(\mathbb{R}^{d}))$, we can similarly obtain the corresponding probability density function of $X_2$, i.e., $q_{X}(\bm{x}) := dF_2(\bm{x})/d\bm{x}$, based on the cumulative distribution function  $F_2(\bm{x}) := Q(X \leq \bm{x}) = Q_{X}((-\infty,\bm{x}])$.
Note that $p_X(\bm{x})$ and $q_X(\bm{x})$ are generally abbreviated to $p(\bm{x})$ and $q(\bm{x})$.
\end{definition}

Based on these definitions, several indices have been introduced in statistics to reflect the dissimilarity between two probability distributions, $p(\bm{x})$ and $q(\bm{x})$.
The Bhattacharyya distance $D_B$ was first proposed by \cite{Bhattacharyya43} as a metric for quantifying dissimilarity:
\begin{align}\label{def_b}
    D_B
    :=& -\ln \int_{\mathbb{R}^{d}} p^{\frac{1}{2}}(\bm{x}) q^{\frac{1}{2}}(\bm{x}) d\bm{x}.
\end{align}
The Chernoff distance $D_C$, an extension of $D_{B}$, was introduced in \cite{Chernoff52}. Here, the square root operator is replaced with an exponent coefficient $s$.
\begin{align}\label{def_c}
    D_C
    :=& -\ln \int_{\mathbb{R}^{d}} p^{s}(\bm{x}) q^{1-s}(\bm{x}) d\bm{x}
\end{align}
The Kullback-Leibler divergence $D_{KL}$, formulated as follows, was proposed in \cite{Kullback51,Kullback59}. Note that it is not a metric because it does not satisfy the metric axiom.
\begin{align}\label{def_kl}
    D_{KL}
    :=& \int_{\mathbb{R}^{d}} \left[ p(\bm{x}) - q(\bm{x}) \right] \ln \left[ \frac{p(\bm{x})}{q(\bm{x})} \right]d\bm{x}
\end{align}
The Hellinger distance $D_H$, as introduced in \cite{Hellinger09}, is defined as follows through the Hellinger integral:
\begin{align}\label{def_h}
    D_H^2
    :=& \frac{1}{2} \int_{\mathbb{R}^{d}} \left[ p^{\frac{1}{2}}(\bm{x}) - q^{\frac{1}{2}}(\bm{x}) \right]^2 d\bm{x}
\end{align}
or
\begin{align}
    D_H
    :=& \frac{1}{\sqrt{2}} \int_{\mathbb{R}^{d}} \left\| p^{\frac{1}{2}}(\bm{x}) - q^{\frac{1}{2}}(\bm{x}) \right\|_2 d\bm{x}.
\end{align}

These measures all give scalar dissimilarities between the two probability distributions $p(\bm{x})$ and $q(\bm{x})$, regardless of the dimensionality of the corresponding random vector.

\section{Main Results}\label{sec:main}

\subsection{Statistical Distance Matrix}

We use the de-trace operation to convert the scalar-valued statistical distances into their matrices. We first focus on the Mahalanobis distance $D_M$  introduced in \cite{Mahalanobis36} for easy understanding because it can be considered a particular case of the Bhattacharyya distance $D_B$. Given two populations with respective mean vectors $\bm{\mu}_1,\bm{\mu}_2 \in \mathbb{R}^{d}$ and covariance matrices $\bm{\Sigma}_1,\bm{\Sigma}_2 \in \mathbb{R}^{d \times d}$, the Mahalanobis distance $D_M$ is written in a quadratic form: 
\begin{align}\label{normal_m}
    D_M
    =& (\bm{\mu}_1 - \bm{\mu}_2)^\intercal \bm{\Sigma}^{-1} (\bm{\mu}_1 - \bm{\mu}_2) \nonumber \\
    =& \tr\left\{(\bm{\mu}_1 - \bm{\mu}_2)(\bm{\mu}_1 - \bm{\mu}_2)^\intercal \bm{\Sigma}^{-1}\right\}
    =   \tr\bm{D}_M,
\end{align}
where $\bm{\Sigma} = \bm{\Sigma}_1 = \bm{\Sigma}_2$.
This quadratic form can be transformed to a trace form, as noted in Eqn. (\ref{normal_m}). By removing the trace, we can obtain the Mahalanobis distance matrix $\bm{D}_M$ in a de-trace form, Eqn. (\ref{normal_m_s}). 

By contrast, the Bhattacharyya distance $D_B$ with corresponding distance matrix $\bm{D}_B$ is defined in a continuous measurable space as per Definition \ref{definition8}. In this paper, we suppose that two $d$-dimensional random vectors $X_1$ and $X_2$ follow two normal distributions, $\mathcal{N}(\bm{\mu}_1, \bm{\Sigma}_1)$ and $\mathcal{N}(\bm{\mu}_2, \bm{\Sigma}_2)$, respectively. Thus, $D_B$ for $X_1$ and $X_2$ is defined as
\begin{align}\label{normal_b}
    D_B
    =& \frac{1}{4}(\bm{\mu}_1 - \bm{\mu}_2)^\intercal (\bm{\Sigma}_1 + \bm{\Sigma}_2 )^{-1} (\bm{\mu}_1 - \bm{\mu}_2) \nonumber \\
    &  + \frac{1}{2} \ln\left[\det\bm{\Sigma}_1^{-\frac{1}{2}} \det\left(\frac{\bm{\Sigma}_1 + \bm{\Sigma}_2}{2}\right) \det\bm{\Sigma}_2^{-\frac{1}{2}}\right] \nonumber \\
    =& \tr \left\{ \frac{1}{4}(\bm{\mu}_1 - \bm{\mu}_2)(\bm{\mu}_1 - \bm{\mu}_2)^\intercal (\bm{\Sigma}_1 + \bm{\Sigma}_2 )^{-1} \right. \nonumber \\
    &  + \left. \frac{1}{2}\left[ \ln\left(\frac{\bm{\Sigma}_1 + \bm{\Sigma}_2}{2}\right) - \ln(\bm{\Sigma}_1^{-\frac{1}{2}}\bm{\Sigma}_2^{-\frac{1}{2}})\right]\right\}
    =   \tr\bm{D}_B.
\end{align}
Appendix \ref{sec:appendix} shows the detailed proof. We find that the first term of Eqn. (\ref{normal_b}) is similar to Eqn. (\ref{normal_m}) and can be transformed to the same trace form. The second term can also be changed into a trace form based on the following equations.
\begin{align} 
    \det \bm{A} \det \bm{B}
    =& \det(\bm{A}\bm{B}), \label{normal_b_p1}\\
    \ln(\det \bm{A})
    =& \tr(\ln \bm{A}), \label{normal_b_p2}\\
    \tr[\ln(\bm{A}\bm{B})]
    =& \tr(\ln \bm{A}) + \tr(\ln \bm{B}) \label{normal_b_p3}
\end{align}
Here, all above equations hold, if and not only if $\bm{A}$ and $\bm{B}$ are two positive definite matrices in $\mathbb{R}^{d \times d}$. Note that Eqn. (\ref{normal_b_p2}) holds by Jacobi’s formula for any complex square matrix where $\ln(\bm{A})$ is defined. Then, by dissolving the trace, the Bhattacharyya distance matrix $\bm{D}_B$ can be written as Eqn. (\ref{normal_b_s}).

The Chernoff distance $D_C$ between the two normal distributions $\mathcal{N}(\bm{\mu}_1, \bm{\Sigma}_1)$ and $\mathcal{N}(\bm{\mu}_2, \bm{\Sigma}_2)$ is defined as Eqn. (\ref{normal_c}), which can be derived by the same way as shown in Appendix \ref{sec:appendix}. After the transformation for obtaining the trace form, the corresponding matrix $\bm{D}_C$ is obtained as Eqn. (\ref{normal_c_s}).
\begin{align}\label{normal_c}
    D_C
    =& \frac{1}{2}s(1-s)(\bm{\mu}_1 - \bm{\mu}_2)^\intercal [(1-s)\bm{\Sigma}_1 + s\bm{\Sigma}_2 ]^{-1} (\bm{\mu}_1 - \bm{\mu}_2) \nonumber \\
    &  + \frac{1}{2} \ln\left\{\det\bm{\Sigma}_1^{s-1} \det\left[(1-s)\bm{\Sigma}_1 + s\bm{\Sigma}_2\right] \det\bm{\Sigma}_2^{-s}\right\} \nonumber \\
    =& \tr \left\{ \frac{1}{2}s(1-s)(\bm{\mu}_1 - \bm{\mu}_2)(\bm{\mu}_1 - \bm{\mu}_2)^\intercal [(1-s)\bm{\Sigma}_1 + s\bm{\Sigma}_2 ]^{-1} \right. \nonumber \\
    &  + \left. \frac{1}{2} \{\ln\left[(1-s)\bm{\Sigma}_1 + s\bm{\Sigma}_2\right] - \ln \left(\bm{\Sigma}_1^{1-s} \bm{\Sigma}_2^{s}\right)\} \right\}
    =   \tr\bm{D}_C,\ \ \ s \in [0,1]
\end{align}
Note that the Chernoff distance $D_C$ with distance matrix $\bm{D}_C$ extends $D_B$ with $\bm{D}_B$, and is more flexible and adaptive due to the exponent coefficient $s$ being adjustable according to computation requirements.

The Kullback-Leibler divergence $D_{KL}$ between $\mathcal{N}(\bm{\mu}_1, \bm{\Sigma}_1)$ and $\mathcal{N}(\bm{\mu}_2, \bm{\Sigma}_2)$ is given by
\begin{align}\label{normal_kl}
    D_{KL}
    =& \frac{1}{2}(\bm{\mu}_1 - \bm{\mu}_2)^\intercal (\bm{\Sigma}_1 + \bm{\Sigma}_2 )^{-1} (\bm{\mu}_1 - \bm{\mu}_2) \nonumber \\
    &  + \frac{1}{2} \tr(\bm{\Sigma}_1^{-1}\bm{\Sigma}_2 + \bm{\Sigma}_2^{-1}\bm{\Sigma}_1 + 2\bm{I}_d) \nonumber \\
    =& \tr \left\{ \frac{1}{2}(\bm{\mu}_1 - \bm{\mu}_2)(\bm{\mu}_1 - \bm{\mu}_2)^\intercal (\bm{\Sigma}_1 + \bm{\Sigma}_2 )^{-1} \right. \nonumber \\
    &  + \left. \frac{1}{2}(\bm{\Sigma}_1^{-1}\bm{\Sigma}_2 + \bm{\Sigma}_2^{-1}\bm{\Sigma}_1 + 2\bm{I}_d) \right\}
    =   \tr\bm{D}_{KL},
\end{align}
where $\bm{I}_d$ is a $d$-dimensional identity matrix. It is easy to write its trace form and obtain the corresponding distance matrix $\bm{D}_{KL}$ as in Eqn. (\ref{normal_kl_s}), where there exists no logarithm operation in the second term.

The Hellinger distance $D_H$ between $\mathcal{N}(\bm{\mu}_1, \bm{\Sigma}_1)$ and $\mathcal{N}(\bm{\mu}_1, \bm{\Sigma}_1)$ is written as
\begin{align}\label{normal_h}
    D_H
    =& \left[1 - \exp(-D_B)\right]^{\frac{1}{2}} \nonumber \\
    =& \left[1 - \exp(-\tr\bm{D}_B)\right]^{\frac{1}{2}}.
\end{align}
It can be considered a function with respect to $\bm{D}_B$, but cannot be changed into a complete trace form. Thus, the Hellinger distance $D_H$ has no distance matrix.

Table \ref{table} summarizes the four types of statistical distance matrices we were able to obtain. 

\renewcommand{\arraystretch}{1.5}
\begin{table}[h]
\begin{tabular}{cc}
  \hline
  Symbol & Parametric formula for definition \\
  \hline \hline
  $\bm{D}_M$ &
  \begin{minipage}{120mm}
  \begin{center}
  \begin{align}\label{normal_m_s}
  (\bm{\mu}_1 - \bm{\mu}_2)(\bm{\mu}_1 - \bm{\mu}_2)^\intercal \bm{\Sigma}^{-1},\ \bm{\Sigma} = \bm{\Sigma}_1 = \bm{\Sigma}_2
  \end{align}
  \end{center}
  \end{minipage} \\
  $\bm{D}_B$ &
  \begin{minipage}{120mm}
  \begin{center}
  \begin{align}\label{normal_b_s}
  \frac{1}{4}(\bm{\mu}_1 - \bm{\mu}_2)(\bm{\mu}_1 - \bm{\mu}_2)^\intercal (\bm{\Sigma}_1 + \bm{\Sigma}_2 )^{-1} + \frac{1}{2}\left[ \ln\left(\frac{\bm{\Sigma}_1 + \bm{\Sigma}_2}{2}\right) - \ln(\bm{\Sigma}_1^{-\frac{1}{2}}\bm{\Sigma}_2^{-\frac{1}{2}})\right]
  \end{align}
  \end{center}
  \end{minipage} \\
  $\bm{D}_C$ &
  \begin{minipage}{120mm}
  \begin{center}
  \begin{align}\label{normal_c_s}
  \frac{1}{2}s(1-s)(\bm{\mu}_1 - \bm{\mu}_2)(\bm{\mu}_1 - \bm{\mu}_2)^\intercal [(1-s)\bm{\Sigma}_1 + s\bm{\Sigma}_2 ]^{-1} \nonumber \\
  + \frac{1}{2} \{\ln\left[(1-s)\bm{\Sigma}_1 + s\bm{\Sigma}_2\right] - \ln \left(\bm{\Sigma}_1^{1-s} \bm{\Sigma}_2^{s}\right)\},\ \ s \in [0,1]
  \end{align}
  \end{center}
  \end{minipage} \\
  $\bm{D}_{KL}$ &
  \begin{minipage}{120mm}
  \begin{center}
  \begin{align}\label{normal_kl_s}
  \frac{1}{2}(\bm{\mu}_1 - \bm{\mu}_2)(\bm{\mu}_1 - \bm{\mu}_2)^\intercal (\bm{\Sigma}_1 + \bm{\Sigma}_2 )^{-1} + \frac{1}{2}(\bm{\Sigma}_1^{-1}\bm{\Sigma}_2 + \bm{\Sigma}_2^{-1}\bm{\Sigma}_1 + 2\bm{I}_d)
  \end{align}
  \end{center}
  \end{minipage} \\ \\ 
  \hline
\end{tabular} 
\caption{Four types of the statistical distance matrices.}\label{table}
\end{table}
\renewcommand{\arraystretch}{1.0} 

\subsection{Hierarchical Clustering for Distance Matrix}

We introduce an ordinary hierarchical clustering \cite{Maimon10, Daniel11}  to cluster the elements of a random vector based on a statistical distance matrix. The input to the hierarchical clustering algorithm is defined as a finite element set $S$  of the random vector $X$ with a distance function $\delta$, which is the map $\delta:S \times S \rightarrow \mathbb{R}$. Here, $\delta(a,b)$ is assigned the element value in a distance matrix $\bm{D}$ of $X$ at location $(a,b)$ and may be zero, where $a,b \in S$ and $\delta(a,a)$ is set to $0$. Given that the set $S$ has $d$ elements, there exist $\binom{d}{2}$ pairwise distances.

The output of the hierarchical clustering algorithm is defined by a dendrogram, which can be considered as a data structure and is expressed as a mathematical graph. A stepwise dendrogram is used in this paper. Given a finite set $S_{0}$ with cardinality $d = |S_{0}|$, a stepwise dendrogram is a list of triples $\langle a_i, b_i, \delta(a_i, b_i) \rangle, i = 0,\ldots,d-2$ with the corresponding node labels $n_i$, where $a_i, b_i \in S_i$. Set $S_0$ is the initial data point. Set $S_{i+1}$ is recursively defined as $(S_i \backslash \{a_i, b_i\}) \cup n_i$. In each step, the new node labeled $n_i$ is formed by joining nodes $a_i$ and $b_i$ at distance $\delta(a_i, b_i)$. The procedure contains $d - 1$ steps, such that the final state is a single node containing all $d$ initial nodes.

The proposed hierarchical clustering algorithm is given in Algorithm \ref{algorithm}.
\begin{algorithm}[H]
    \caption{Hierarchical Clustering Algorithm for Distance Matrix}
    \label{algorithm}
    \begin{algorithmic}[1]

        \Inputs{Node labels: $S_0$ \\ Distance function: $\delta$} \vspace{1mm}
        \Initialize{Number of input nodes: $d \leftarrow |S|$ \\ Stepwise dendrogram: $L \leftarrow \varnothing$}

        \For{$i = 0$ \textbf{to} $d - 2$}
            \State $(a_i,b_i) \leftarrow \argmin_{S_i \times S_i \backslash \Delta_i}$, $\Delta_i$ denotes diagonal elements in $S_i \times S_i$.
            \State Append triple $\langle a_i, b_i, d(a_i,b_i) \rangle$ to $L$
            \State $S_i \leftarrow S_i \backslash \{a_i, b_i\}$
            \State Create a new node label $n_i \notin S_i$
            \State Update $\delta$ for all $x \in S_i$ by
            \begin{align}
                \hspace{6mm}\delta(n_i,x) = \delta(x,n_i) := f(\delta(a_i,x),\delta(b_i,x)) \nonumber
            \end{align}
            \State $S_{i+1} \leftarrow S_i \cup \{n_i\}$
        \EndFor

        \Outputs{Stepwise dendrogram: $L$}
  \end{algorithmic}
\end{algorithm}
Here, the agglomerative formula for updating $\delta$ is defined as
\begin{align}
    f(d(a_i,x),d(b_i,x))
    :=& \max(d(a_i,x),d(b_i,x)).
\end{align}
Given a cut-off threshold, this algorithm can provide stable and reliable clustering results for the elements of a random vector.

\section{Computed Examples}\label{sec:simulations}

We use the CIFAR-10 dataset \cite{CIFAR} to test the effects of the statistical distance matrices. This dataset contains 60,000 $32 \times 32$ color images in 10 different classes. To simplify the calculation and obtain distinguishable results, as shown in Figure ref{fig:example}, we calculated only the distance matrices between airplanes and dogs, birds and dogs, cats and dogs, such that the similarities between every two objects could range from weak to strong. 

\begin{figure}[tb]
\begin{center}
\begin{tabular}{llllll}

\begin{minipage}{22mm}
\begin{center}
Airplane
\end{center}
\end{minipage}&

\begin{minipage}{22mm}
\begin{center}
\includegraphics[width=22mm]{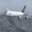}\\
\end{center}
\end{minipage}&

\begin{minipage}{22mm}
\begin{center}
\includegraphics[width=22mm]{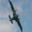}\\
\end{center}
\end{minipage}&

\begin{minipage}{22mm}
\begin{center}
\includegraphics[width=22mm]{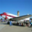}\\
\end{center}
\end{minipage}&

\begin{minipage}{22mm}
\begin{center}
\includegraphics[width=22mm]{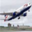}\\
\end{center}
\end{minipage}&

\begin{minipage}{10mm}
\begin{center}
$\cdots$
\end{center}
\end{minipage} \\ \\

\begin{minipage}{22mm}
\begin{center}
Bird
\end{center}
\end{minipage}&

\begin{minipage}{22mm}
\begin{center}
\includegraphics[width=22mm]{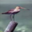}\\
\end{center}
\end{minipage}&

\begin{minipage}{22mm}
\begin{center}
\includegraphics[width=22mm]{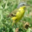}\\
\end{center}
\end{minipage}&

\begin{minipage}{22mm}
\begin{center}
\includegraphics[width=22mm]{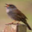}\\
\end{center}
\end{minipage}&

\begin{minipage}{22mm}
\begin{center}
\includegraphics[width=22mm]{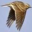}\\
\end{center}
\end{minipage}&

\begin{minipage}{10mm}
\begin{center}
$\cdots$
\end{center}
\end{minipage} \\ \\

\begin{minipage}{22mm}
\begin{center}
Cat
\end{center}
\end{minipage}&

\begin{minipage}{22mm}
\begin{center}
\includegraphics[width=22mm]{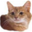}\\
\end{center}
\end{minipage}&

\begin{minipage}{22mm}
\begin{center}
\includegraphics[width=22mm]{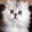}\\
\end{center}
\end{minipage}&

\begin{minipage}{22mm}
\begin{center}
\includegraphics[width=22mm]{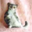}\\
\end{center}
\end{minipage}&

\begin{minipage}{22mm}
\begin{center}
\includegraphics[width=22mm]{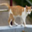}\\
\end{center}
\end{minipage}&

\begin{minipage}{10mm}
\begin{center}
$\cdots$
\end{center}
\end{minipage} \\ \\

\begin{minipage}{22mm}
\begin{center}
Dog
\end{center}
\end{minipage}&

\begin{minipage}{22mm}
\begin{center}
\includegraphics[width=22mm]{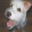}\\
\end{center}
\end{minipage}&

\begin{minipage}{22mm}
\begin{center}
\includegraphics[width=22mm]{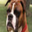}\\
\end{center}
\end{minipage}&

\begin{minipage}{22mm}
\begin{center}
\includegraphics[width=22mm]{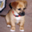}\\
\end{center}
\end{minipage}&

\begin{minipage}{22mm}
\begin{center}
\includegraphics[width=22mm]{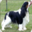}\\
\end{center}
\end{minipage}&

\begin{minipage}{10mm}
\begin{center}
$\cdots$
\end{center}
\end{minipage}

\end{tabular}
\end{center}
\caption{Image examples of airplanes, birds, cats, and dogs in the CIFAR-10 dataset.}
\label{fig:example}
\end{figure}

Given a value space $U = [0,1]$ for an image pixel, an image $\bm{A} := [a_{ij}] \in U^{d \times d},\ i,j = 1,\ldots,d$ is re-formed as $\bm{a} := [\acute{a}_t] = \ve \bm{A}\ \in U^{d^2 \times 1},\ t = 1,\ldots,d^2$ through matrix vectorization $\ve(\cdot)$. Let image sets $\{\bm{a}_k\}$ and $\{\bm{b}_k\}$, $k = 1,\ldots,N$  of two classes be regarded as two populations with $d$-dimensional random vectors $X_1$ and $X_2$, which respectively follow the two normal distributions $\mathcal{N}(\bm{\mu}_1, \bm{\Sigma}_1)$ and $\mathcal{N}(\bm{\mu}_2, \bm{\Sigma}_2)$, where
\begin{align}\label{normal_mean}
    \bm{\mu}_1
    :=& \mathbb{E}(\bm{a}_k) = \frac{1}{N} \sum_{k=1}^{N}\bm{a}_k,\ \
    \bm{\Sigma}_1
    :=   \cov(\bm{a}_k) = \frac{1}{N-1} \sum_{k=1}^{N}[(\bm{a}_k - \bm{\mu}_1)(\bm{a}_k - \bm{\mu}_1)^\intercal], \nonumber \\
    \bm{\mu}_2
    :=& \mathbb{E}(\bm{b}_k) = \frac{1}{N} \sum_{k=1}^{N}\bm{b}_k,\ \
    \bm{\Sigma}_2
    :=   \cov(\bm{b}_k) = \frac{1}{N-1} \sum_{k=1}^{N}[(\bm{b}_k - \bm{\mu}_2)(\bm{b}_k - \bm{\mu}_2)^\intercal].
\end{align}

\begin{figure}[H]
\begin{center}
\begin{tabular}{llll}

&

\begin{minipage}{40mm}
\begin{center}
Airplane/Dog
\end{center}
\end{minipage}&

\begin{minipage}{40mm}
\begin{center}
Bird/Dog
\end{center}
\end{minipage}&

\begin{minipage}{40mm}
\begin{center}
Cat/Dog
\end{center}
\end{minipage}\\ \\

\begin{minipage}{5mm}
\begin{center}
$\bm{D}_M$
\end{center}
\end{minipage}&

\begin{minipage}{30mm}
\begin{center}
\includegraphics[width=40mm]{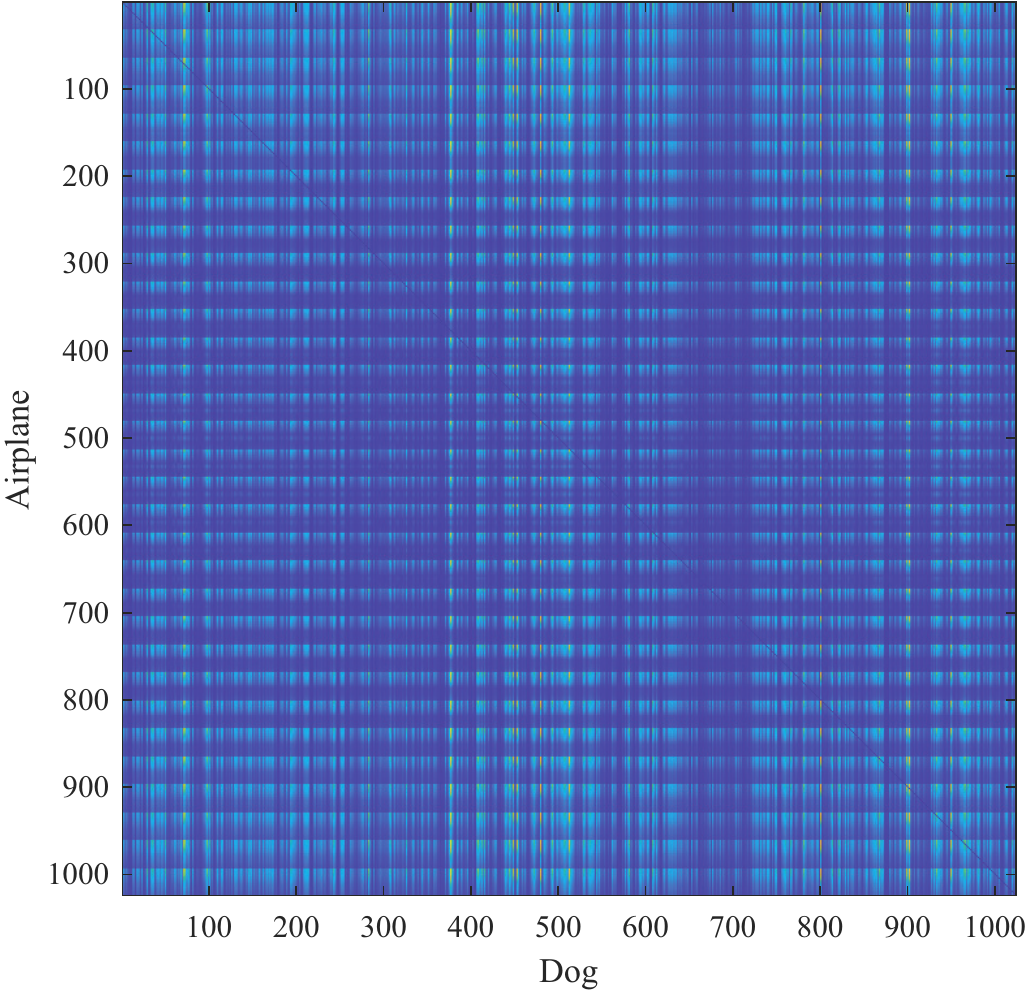}\\
\end{center}
\end{minipage}&

\begin{minipage}{30mm}
\begin{center}
\includegraphics[width=40mm]{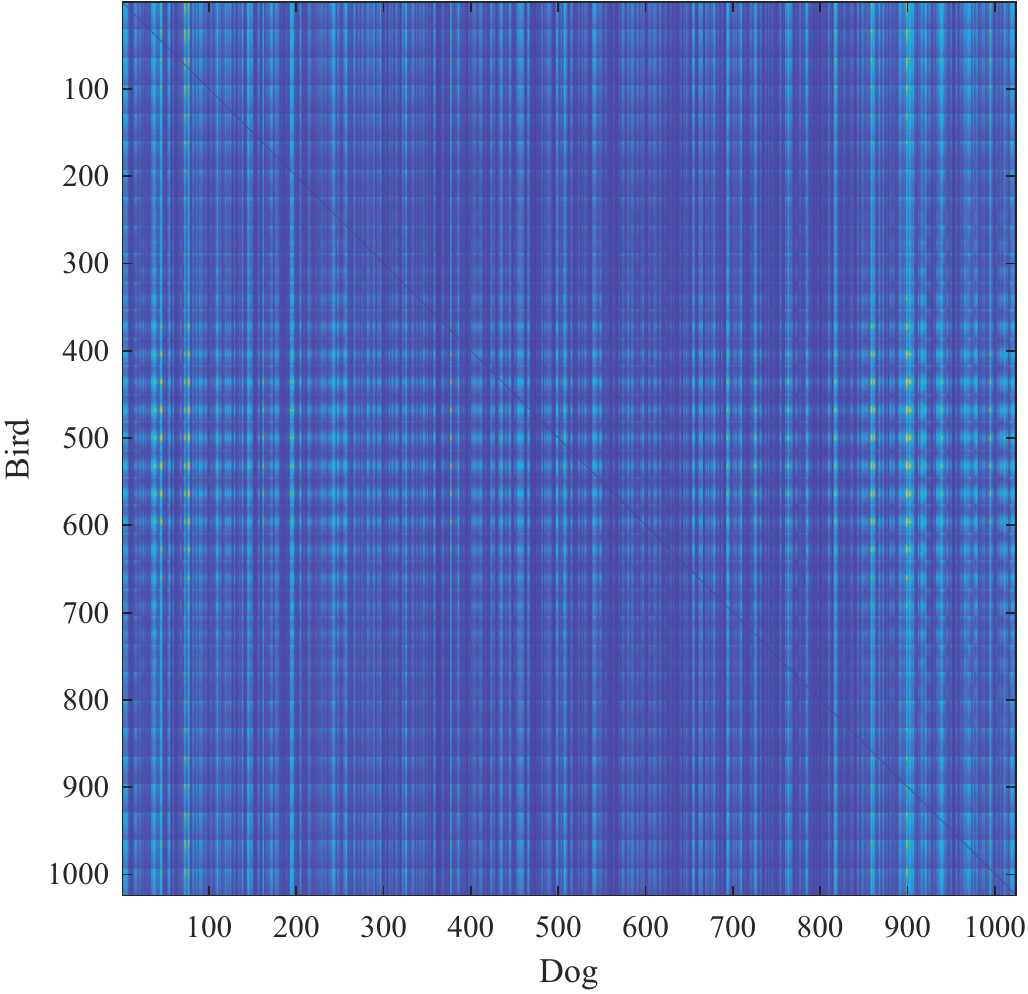}\\
\end{center}
\end{minipage}&

\begin{minipage}{30mm}
\begin{center}
\includegraphics[width=40mm]{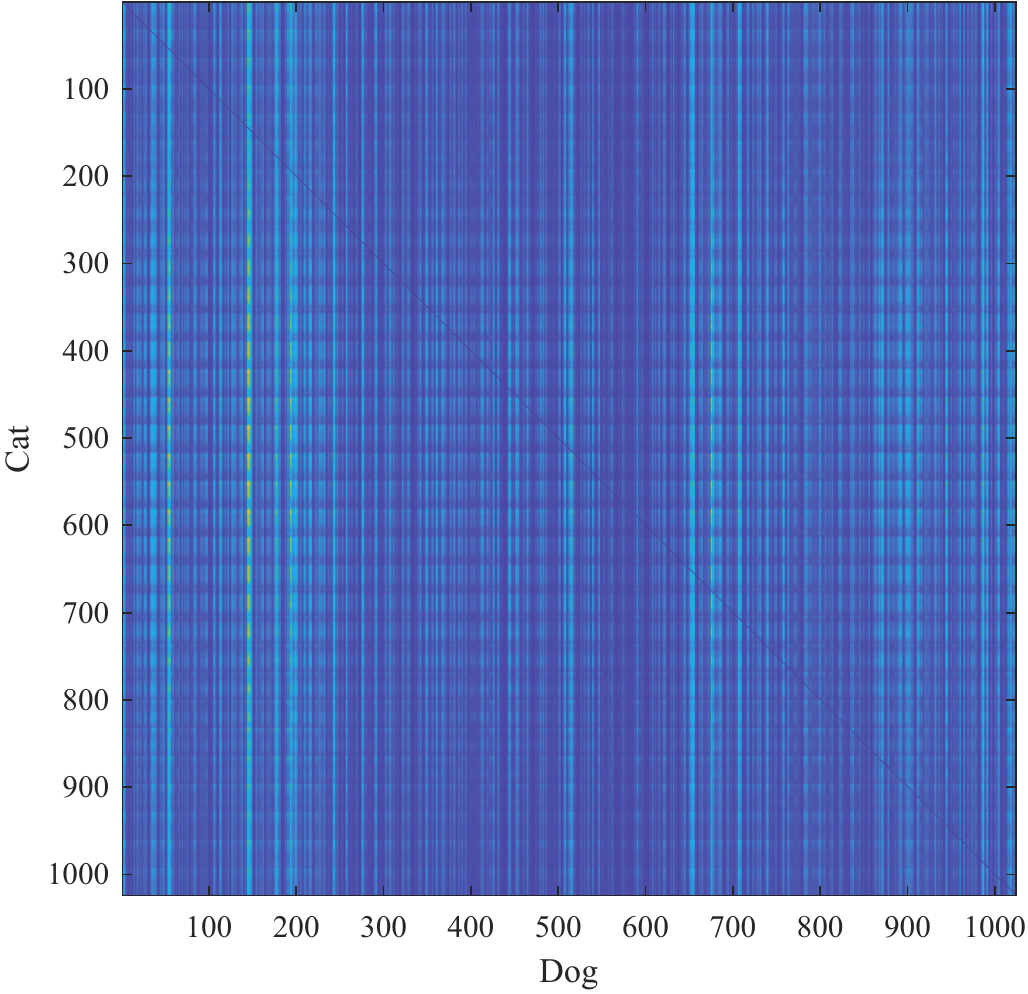}\\
\end{center}
\end{minipage}\\ \\

\begin{minipage}{5mm}
\begin{center}
$\bm{D}_{KL}$
\end{center}
\end{minipage}&

\begin{minipage}{30mm}
\begin{center}
\includegraphics[width=40mm]{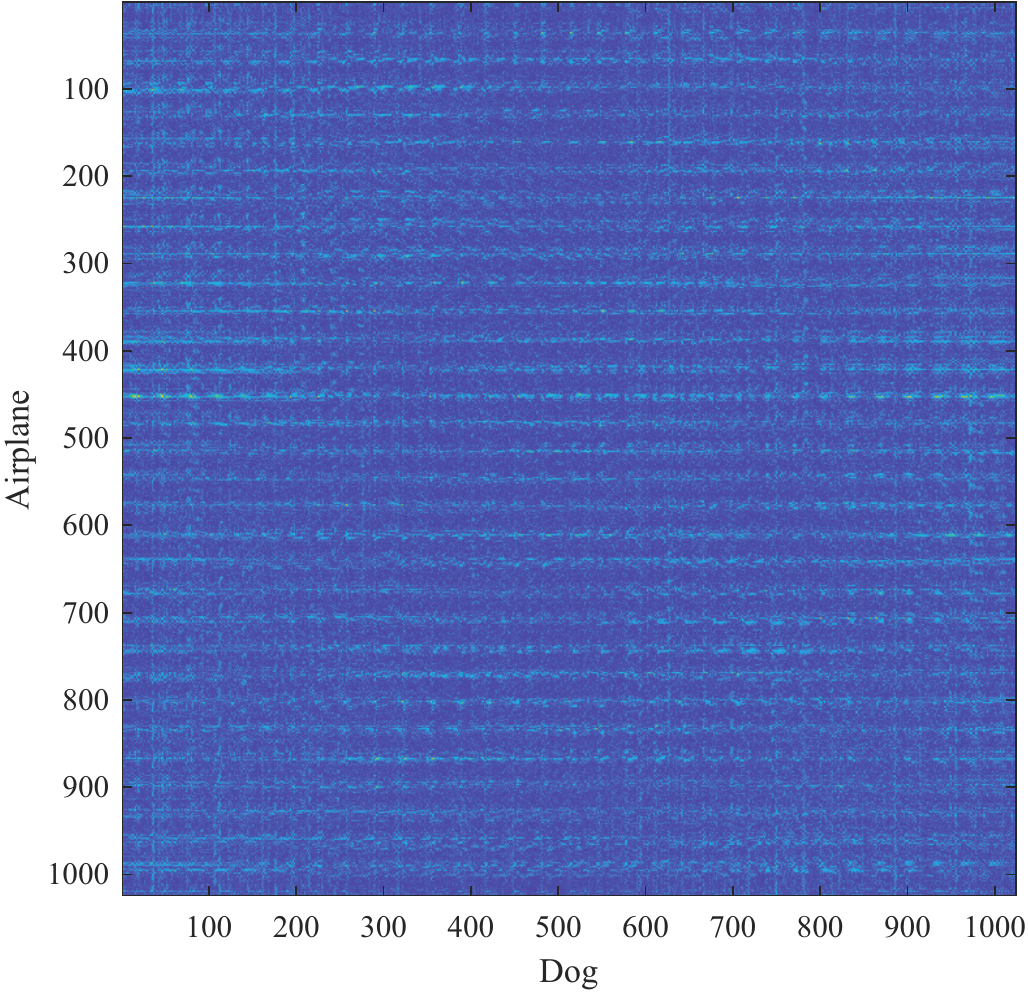}\\
\end{center}
\end{minipage}&

\begin{minipage}{30mm}
\begin{center}
\includegraphics[width=40mm]{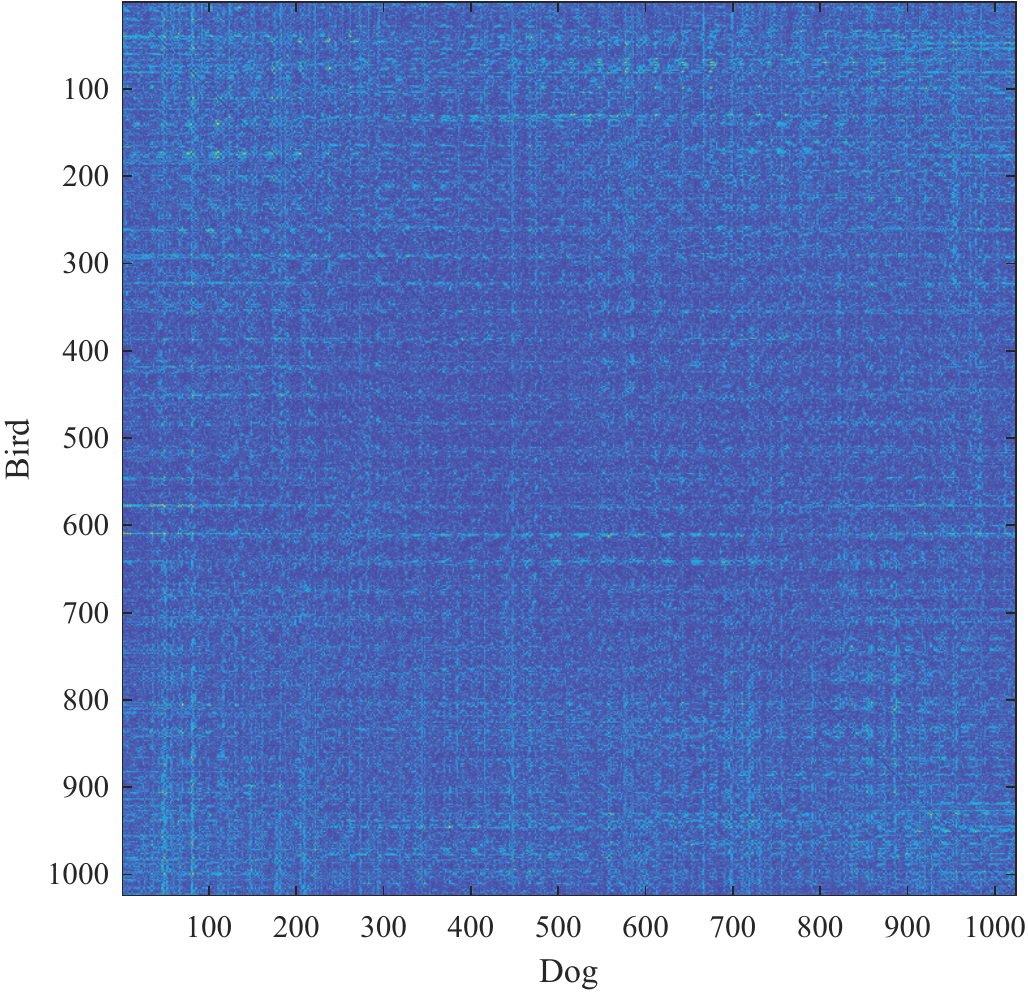}\\
\end{center}
\end{minipage}&

\begin{minipage}{30mm}
\begin{center}
\includegraphics[width=40mm]{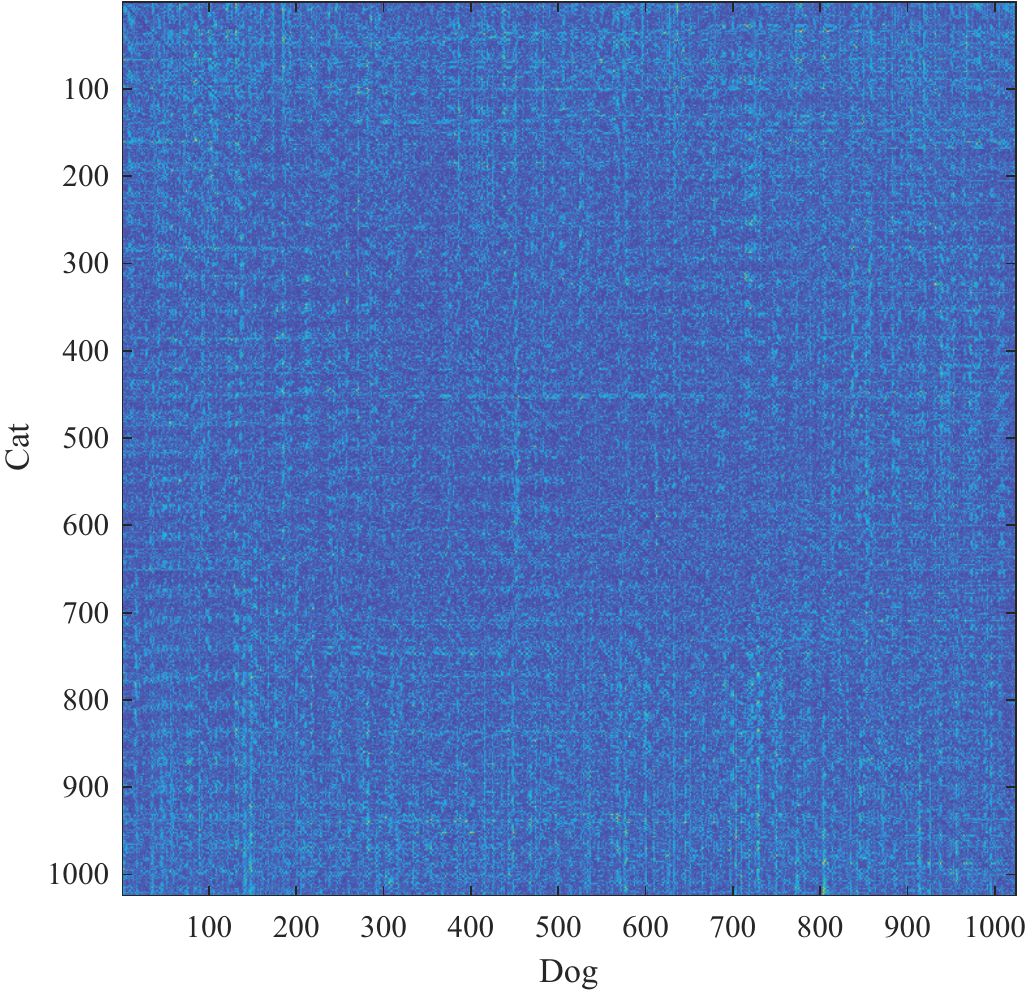}\\
\end{center}
\end{minipage}\\ \\

\begin{minipage}{5mm}
\begin{center}
$\bm{D}_B$
\end{center}
\end{minipage}&

\begin{minipage}{30mm}
\begin{center}
\includegraphics[width=40mm]{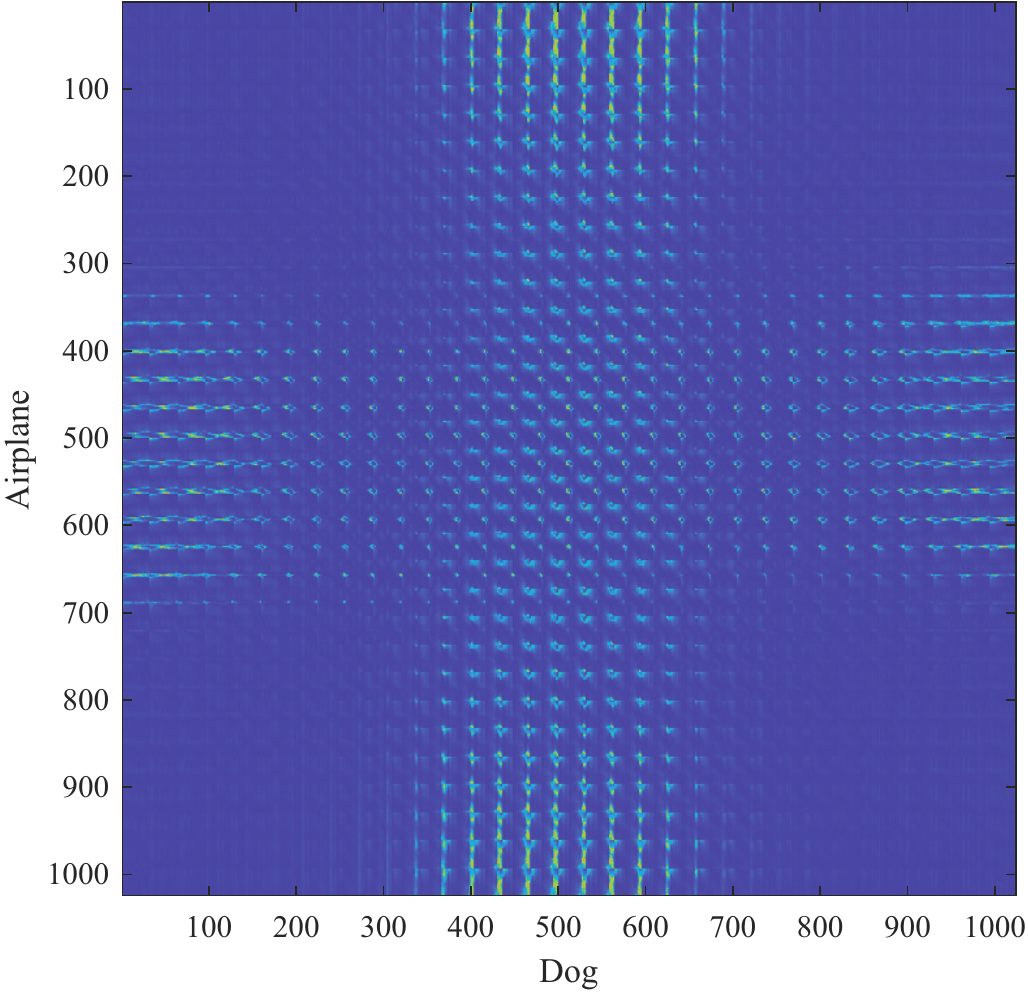}\\
\end{center}
\end{minipage}&

\begin{minipage}{30mm}
\begin{center}
\includegraphics[width=40mm]{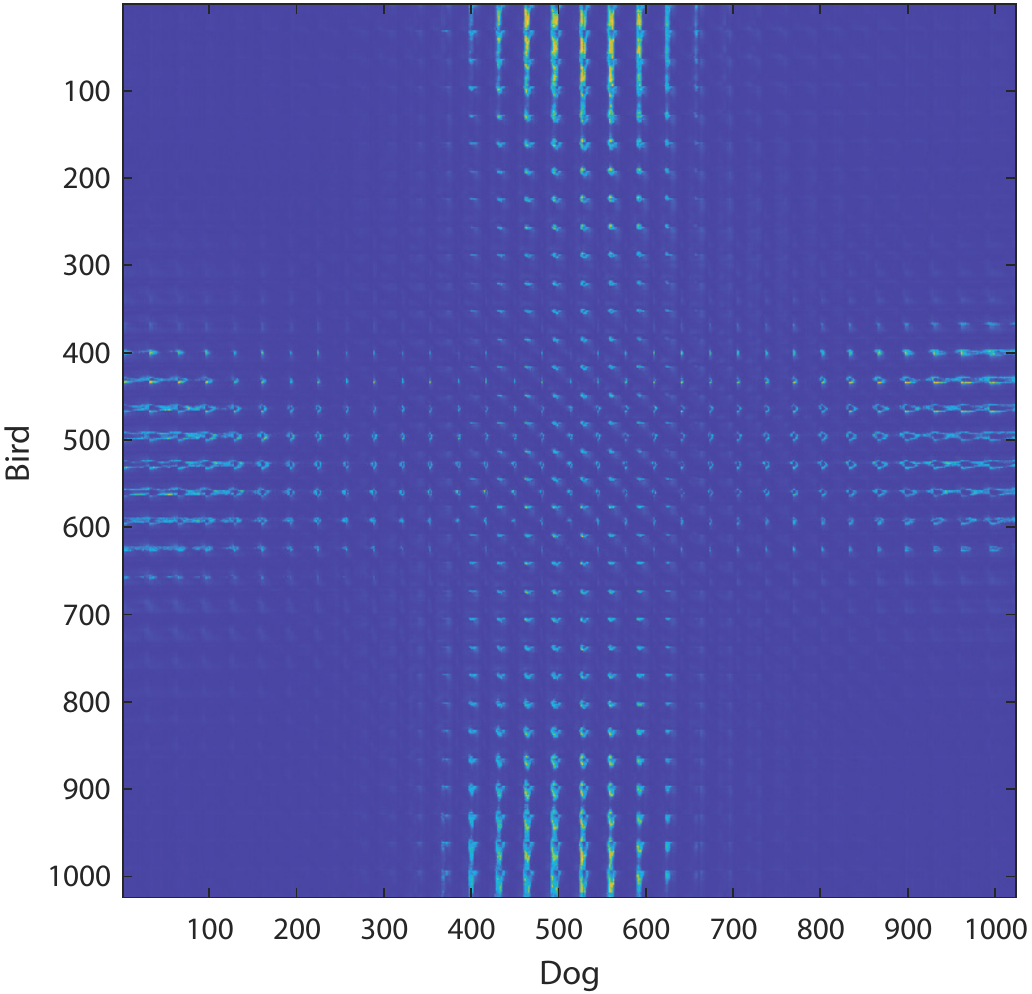}\\
\end{center}
\end{minipage}&

\begin{minipage}{30mm}
\begin{center}
\includegraphics[width=40mm]{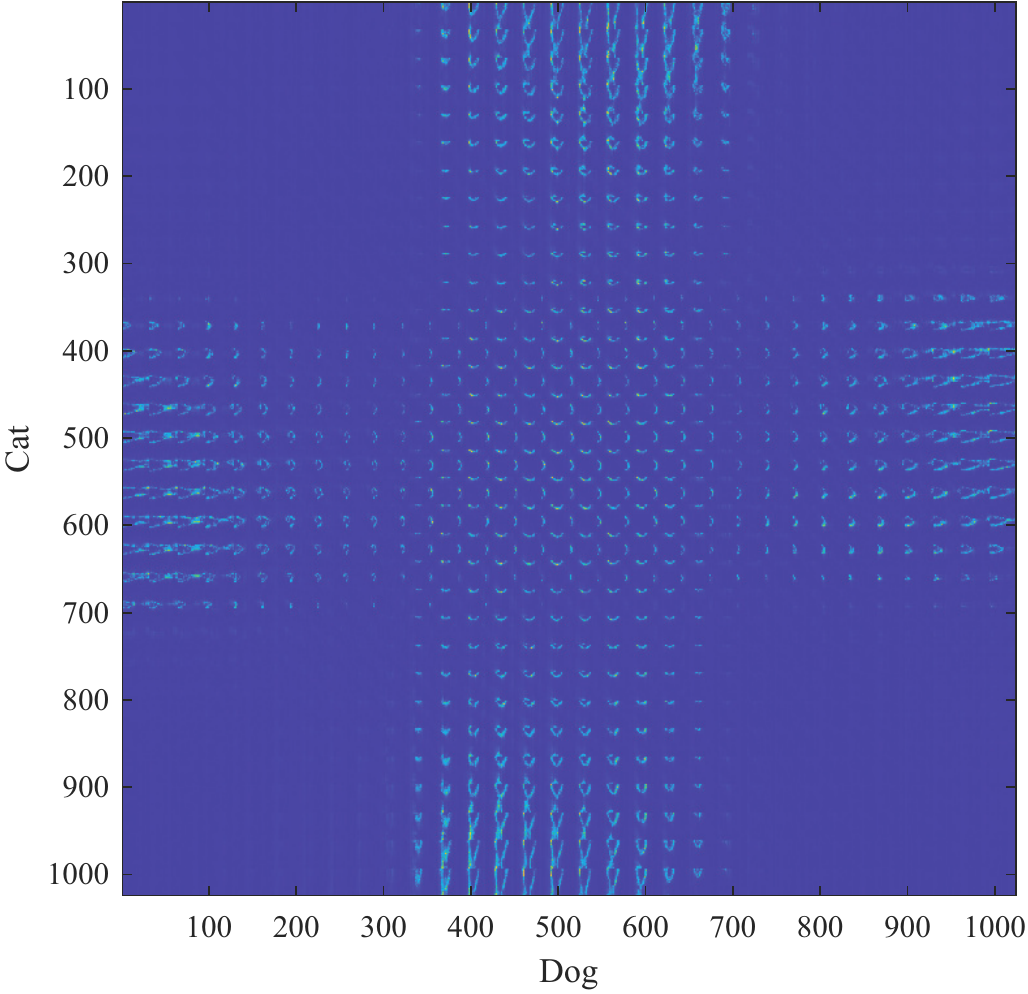}\\
\end{center}
\end{minipage}\\ \\

\begin{minipage}{5mm}
\begin{center}
$\bm{D}_C$
\end{center}
\end{minipage}&

\begin{minipage}{30mm}
\begin{center}
\includegraphics[width=40mm]{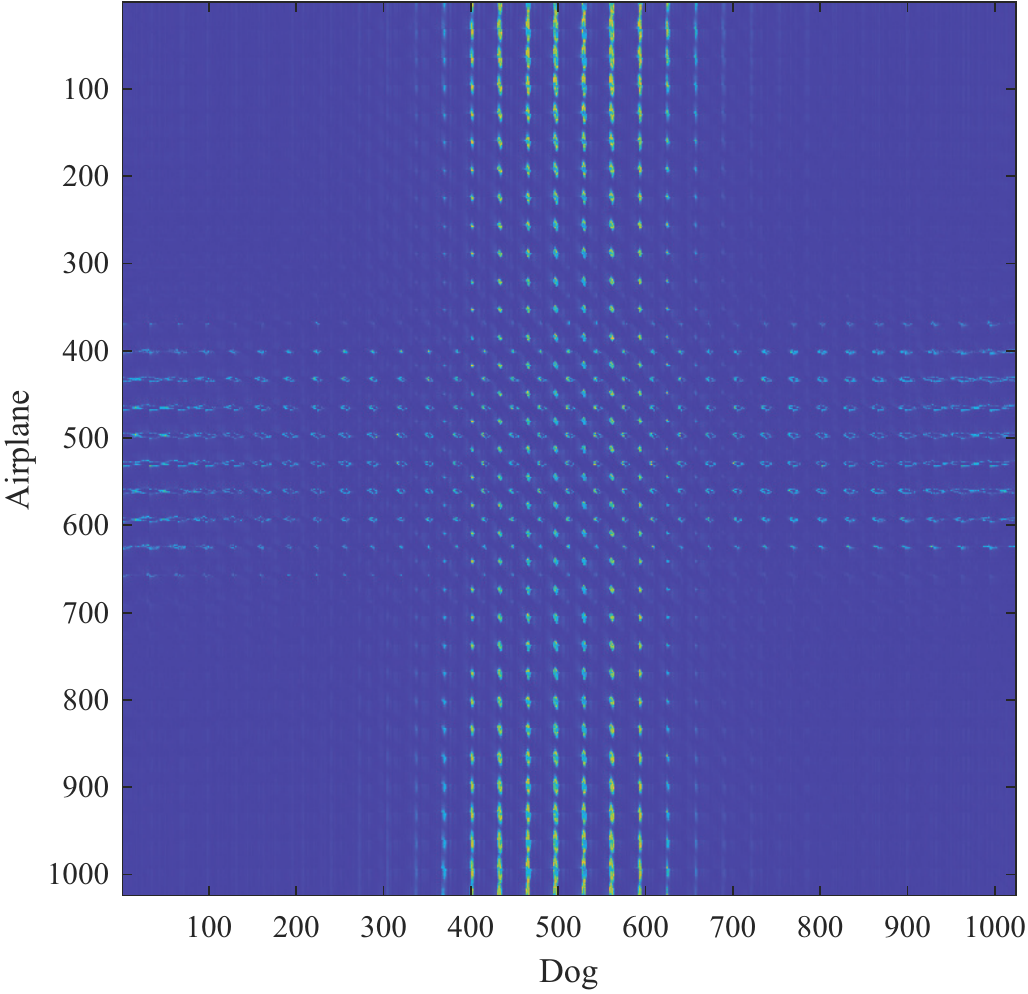}\\
\end{center}
\end{minipage}&

\begin{minipage}{30mm}
\begin{center}
\includegraphics[width=40mm]{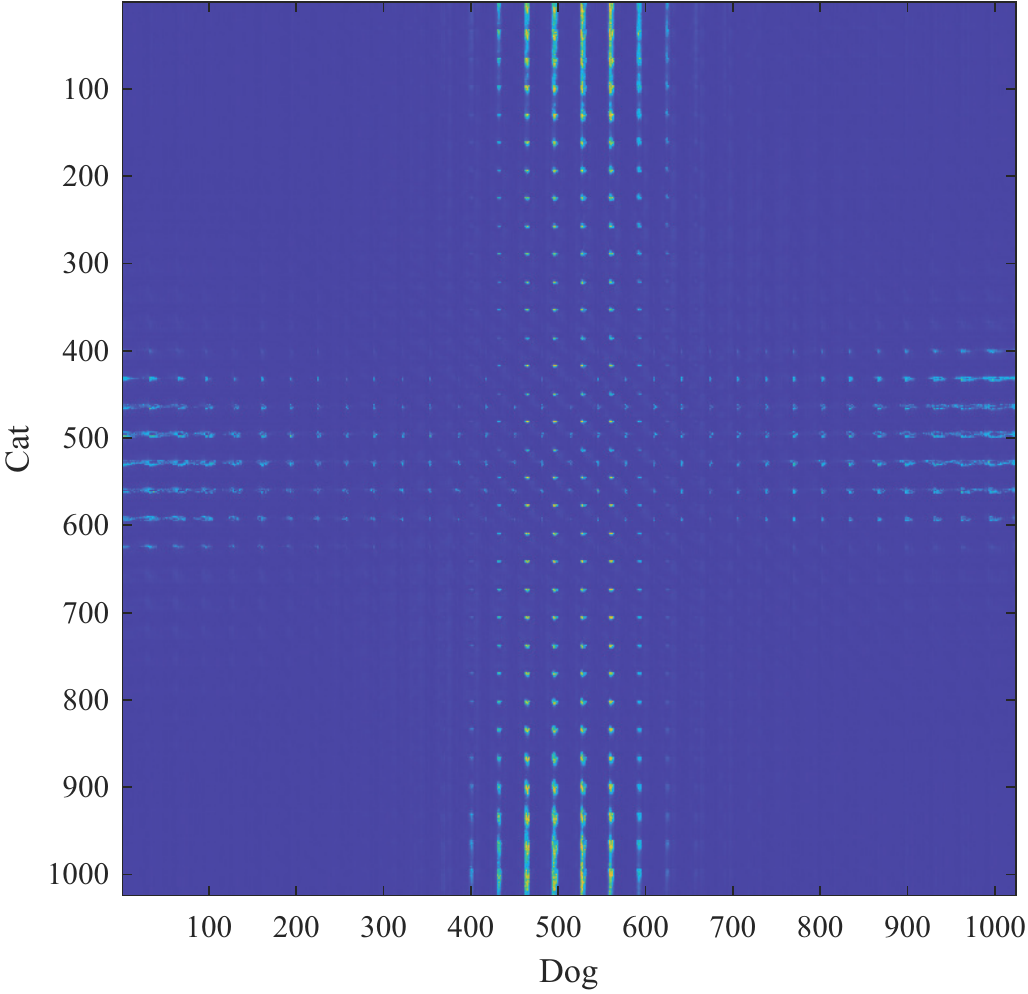}\\
\end{center}
\end{minipage}&

\begin{minipage}{30mm}
\begin{center}
\includegraphics[width=40mm]{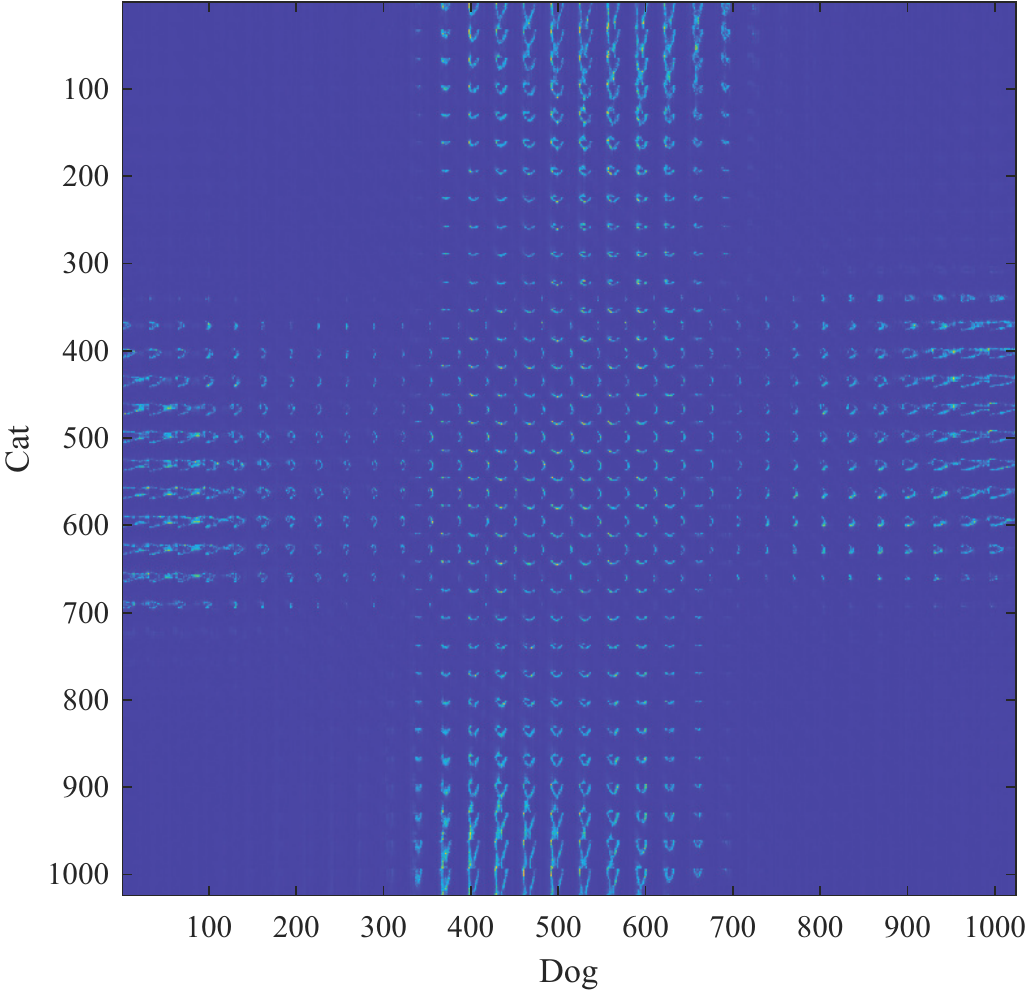}\\
\end{center}
\end{minipage}

\end{tabular}
\end{center}
\caption{Statistical distance matrices $\bm{D}_M$, $\bm{D}_{KL}$, $\bm{D}_B$, and $\bm{D}_C$ for the cases of airplanes and dogs, birds and dogs, cats and dogs, respectively.}
\label{fig:dis_matrix}
\end{figure}

In the experiments, $\{\bm{a}_k\}$ and $\{\bm{b}_k\}$ were set to the image sets of airplanes and dogs, birds and dogs, and cats and dogs in order. Note that $N = 5000$ because of only the training images are used for each class. By calculating and substituting their mean vectors $\bm{\mu}_1$ and $\bm{\mu}_2$ with covariance matrices $\bm{\Sigma}_1$ and $\bm{\Sigma}_2$ from rom Eqn. (\ref{normal_mean}) into Eqn. (\ref{normal_m_s})-(\ref{normal_kl_s}), we can obtain the four statistical distance matrices $\bm{D}_M$, $\bm{D}_{KL}$, $\bm{D}_B$, and $\bm{D}_C$ for all of three cases, where each $\bm{D}:=[\delta_{uv}] \in \mathbb{R}^{d^2 \times d^2}$, $u,v = 1,\ldots,d^2$. 

\begin{figure}[H]
\begin{center}
\begin{tabular}{llll}

&

\begin{minipage}{40mm}
\begin{center}
Airplane/Dog
\end{center}
\end{minipage}&

\begin{minipage}{40mm}
\begin{center}
Bird/Dog
\end{center}
\end{minipage}&

\begin{minipage}{40mm}
\begin{center}
Cat/Dog
\end{center}
\end{minipage}\\ \\

\begin{minipage}{5mm}
\begin{center}
$\bm{\Phi}_M$
\end{center}
\end{minipage}&

\begin{minipage}{40mm}
\begin{center}
\includegraphics[width=40mm]{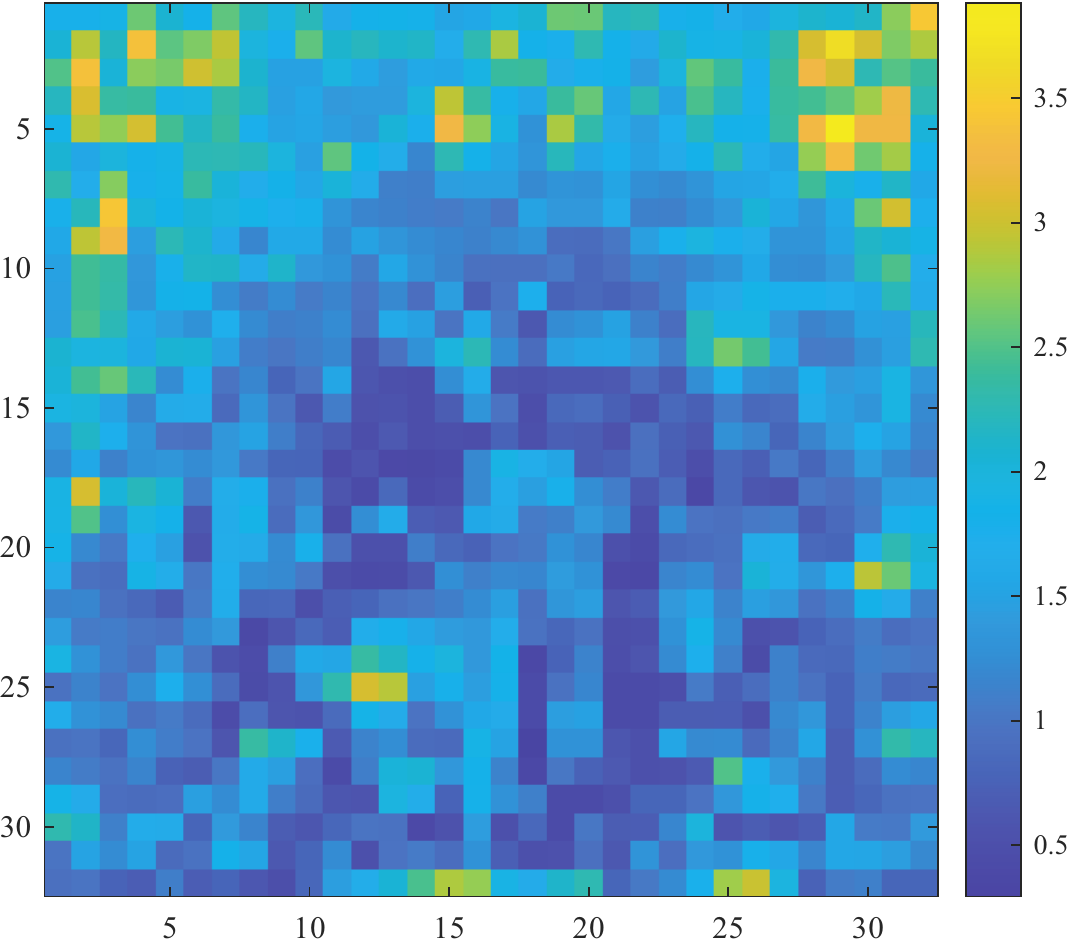}\\
\end{center}
\end{minipage}&

\begin{minipage}{40mm}
\begin{center}
\includegraphics[width=40mm]{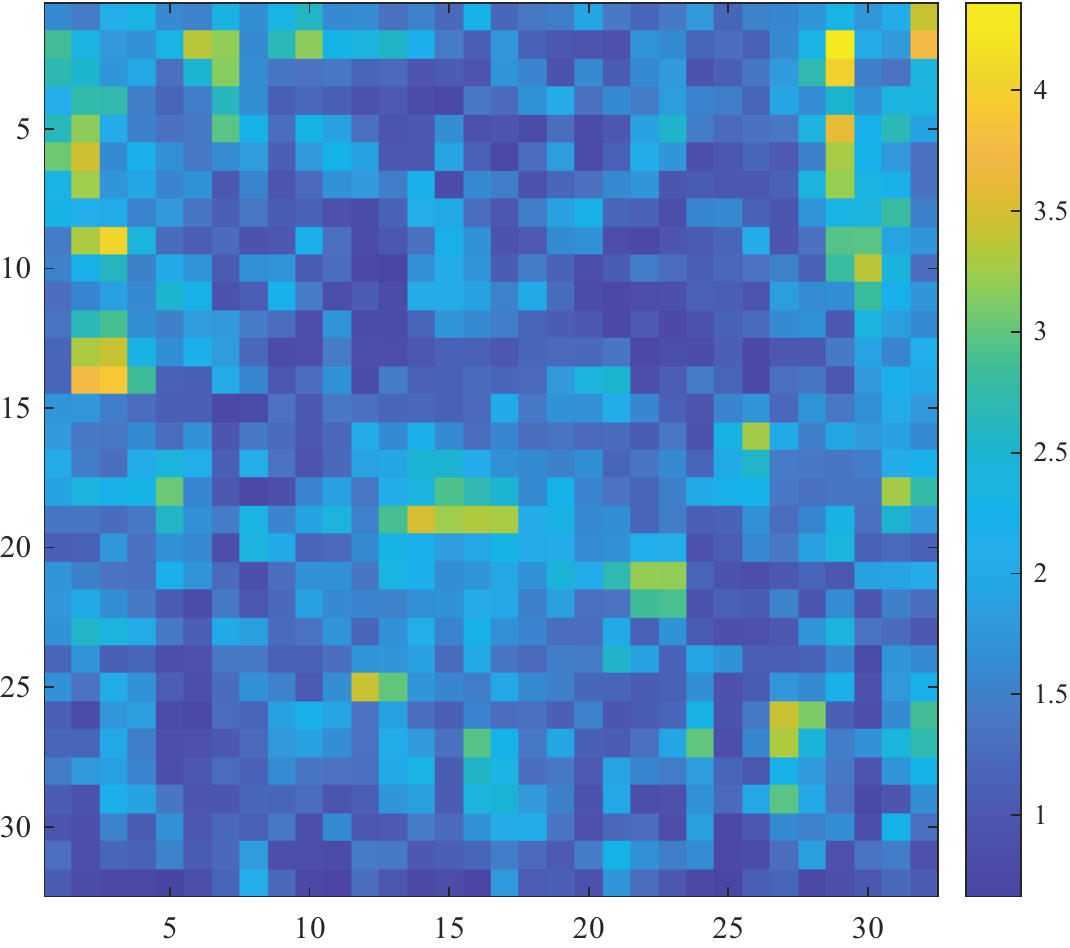}\\
\end{center}
\end{minipage}&

\begin{minipage}{40mm}
\begin{center}
\includegraphics[width=40mm]{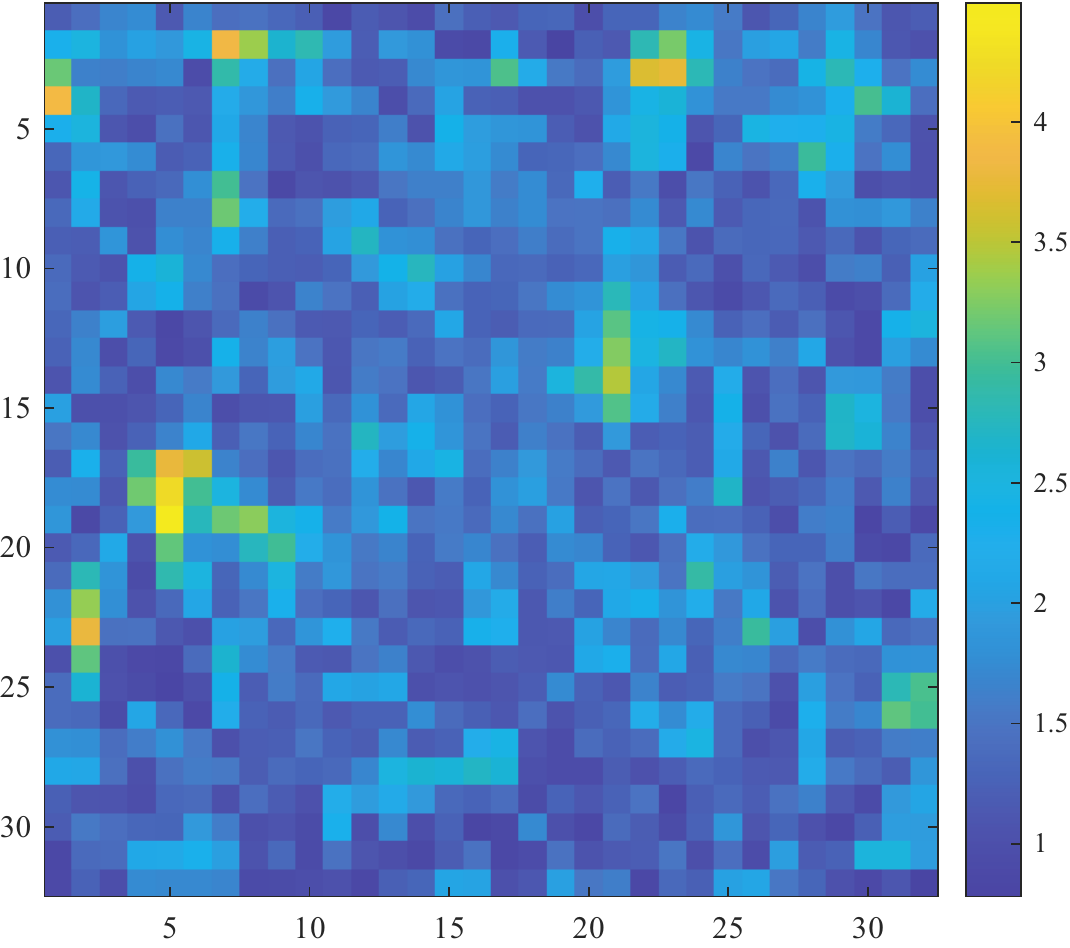}\\
\end{center}
\end{minipage}\\ \\

\begin{minipage}{5mm}
\begin{center}
$\bm{\Phi}_{KL}$
\end{center}
\end{minipage}&

\begin{minipage}{40mm}
\begin{center}
\includegraphics[width=40mm]{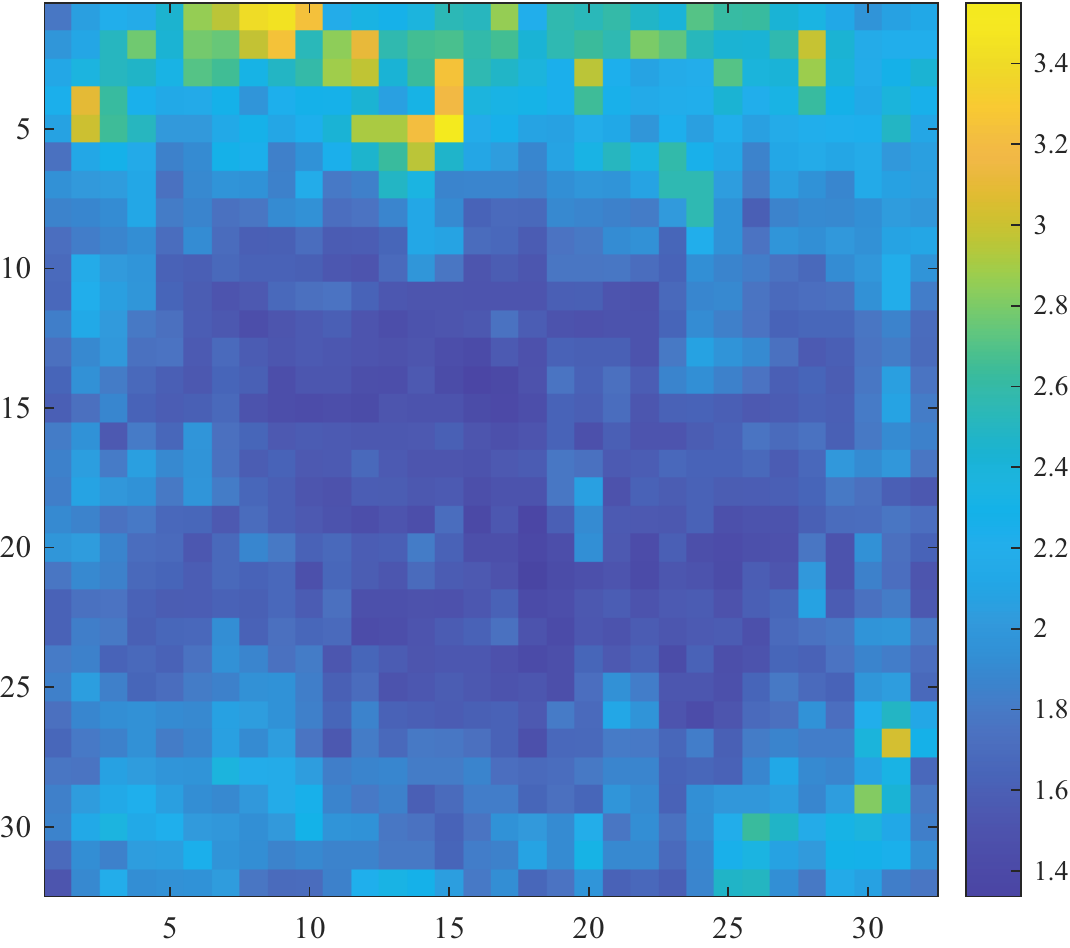}\\
\end{center}
\end{minipage}&

\begin{minipage}{40mm}
\begin{center}
\includegraphics[width=40mm]{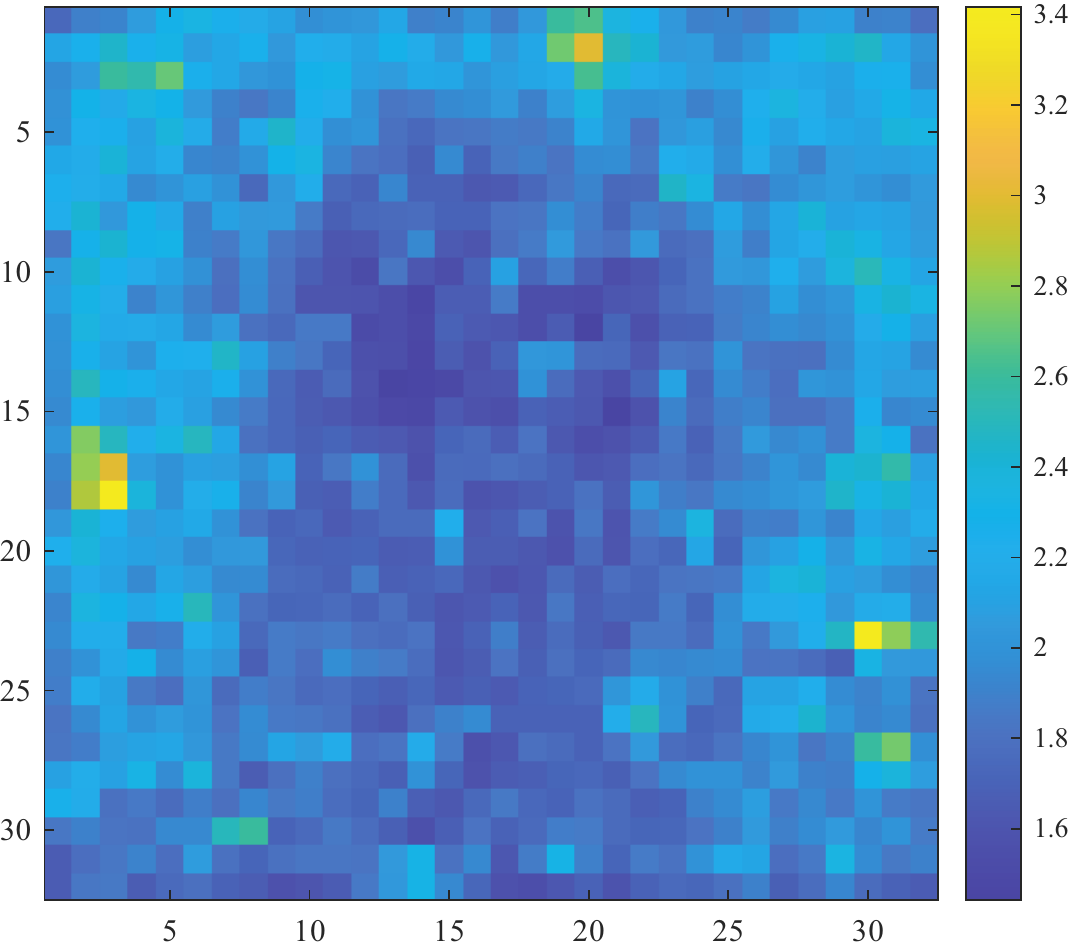}\\
\end{center}
\end{minipage}&

\begin{minipage}{40mm}
\begin{center}
\includegraphics[width=40mm]{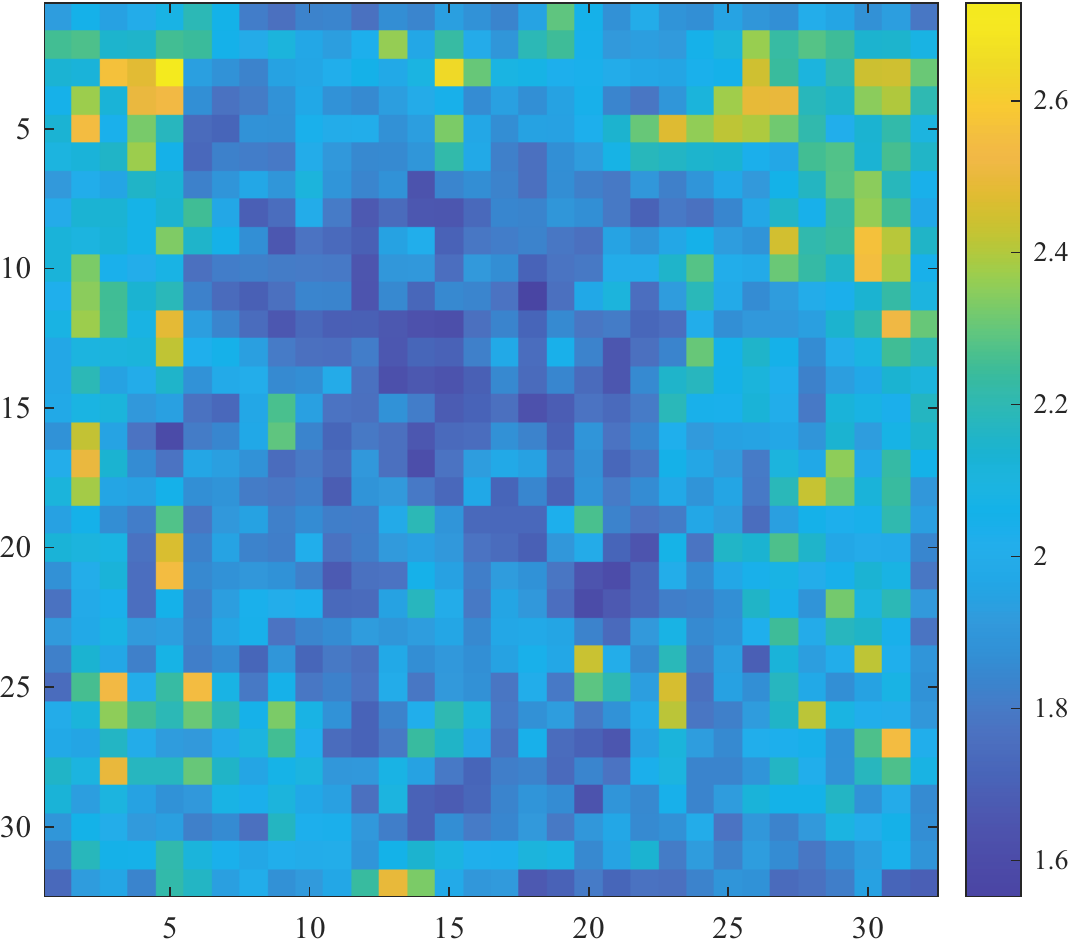}\\
\end{center}
\end{minipage}\\ \\

\begin{minipage}{5mm}
\begin{center}
$\bm{\Phi}_B$
\end{center}
\end{minipage}&

\begin{minipage}{40mm}
\begin{center}
\includegraphics[width=40mm]{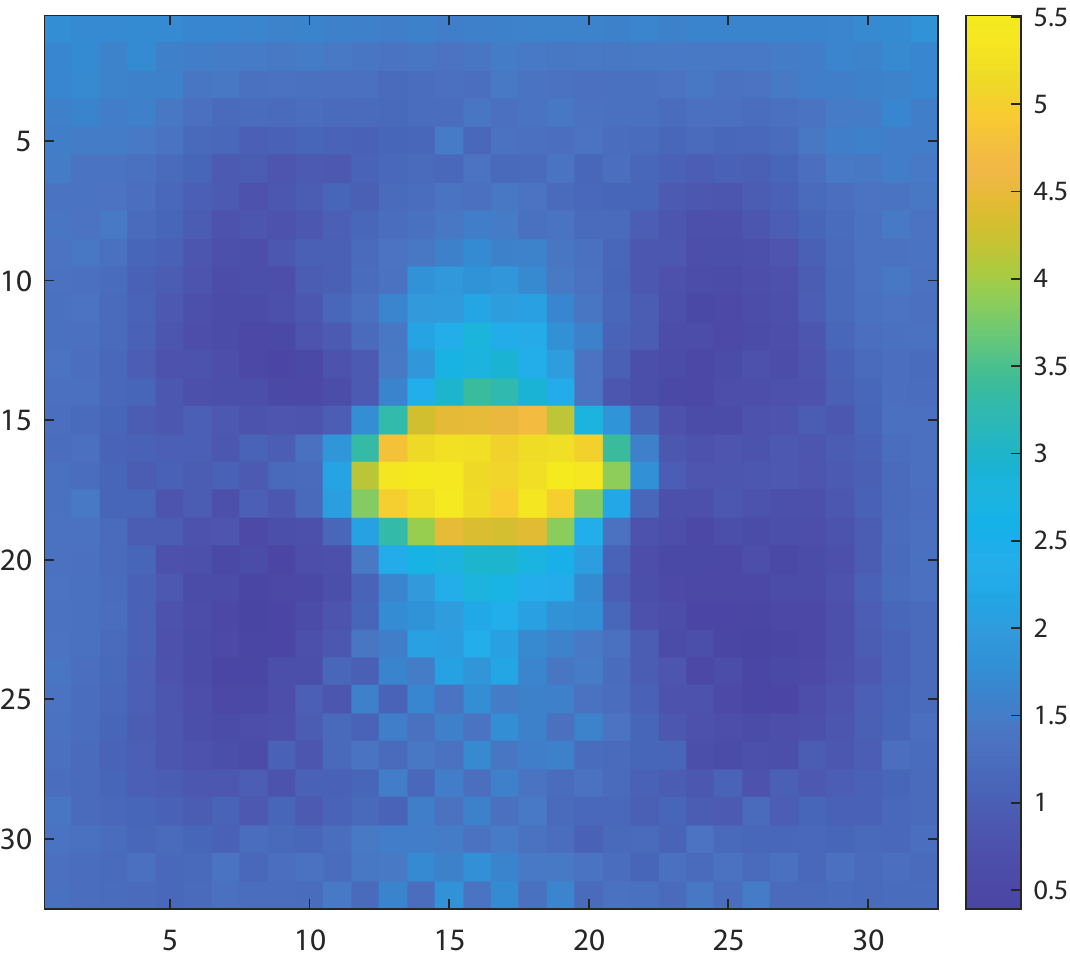}\\
\end{center}
\end{minipage}&

\begin{minipage}{40mm}
\begin{center}
\includegraphics[width=40mm]{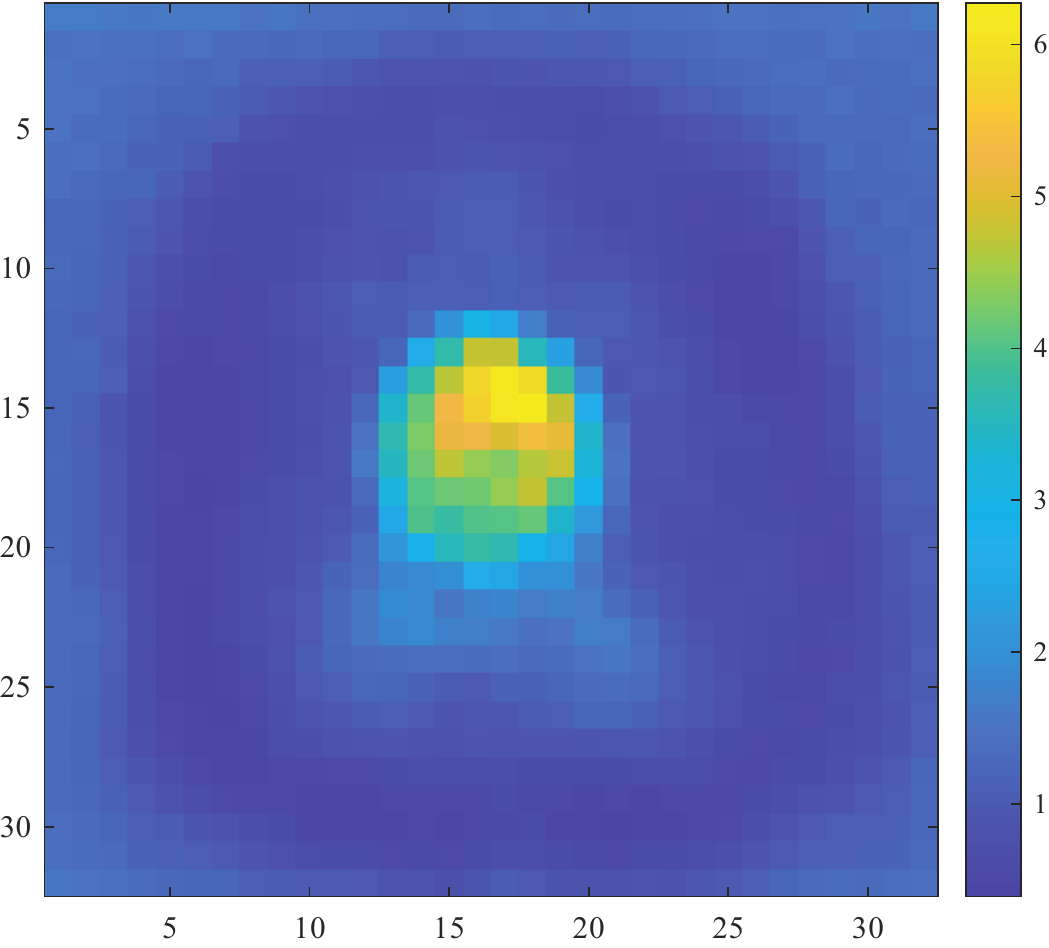}\\
\end{center}
\end{minipage}&

\begin{minipage}{40mm}
\begin{center}
\includegraphics[width=40mm]{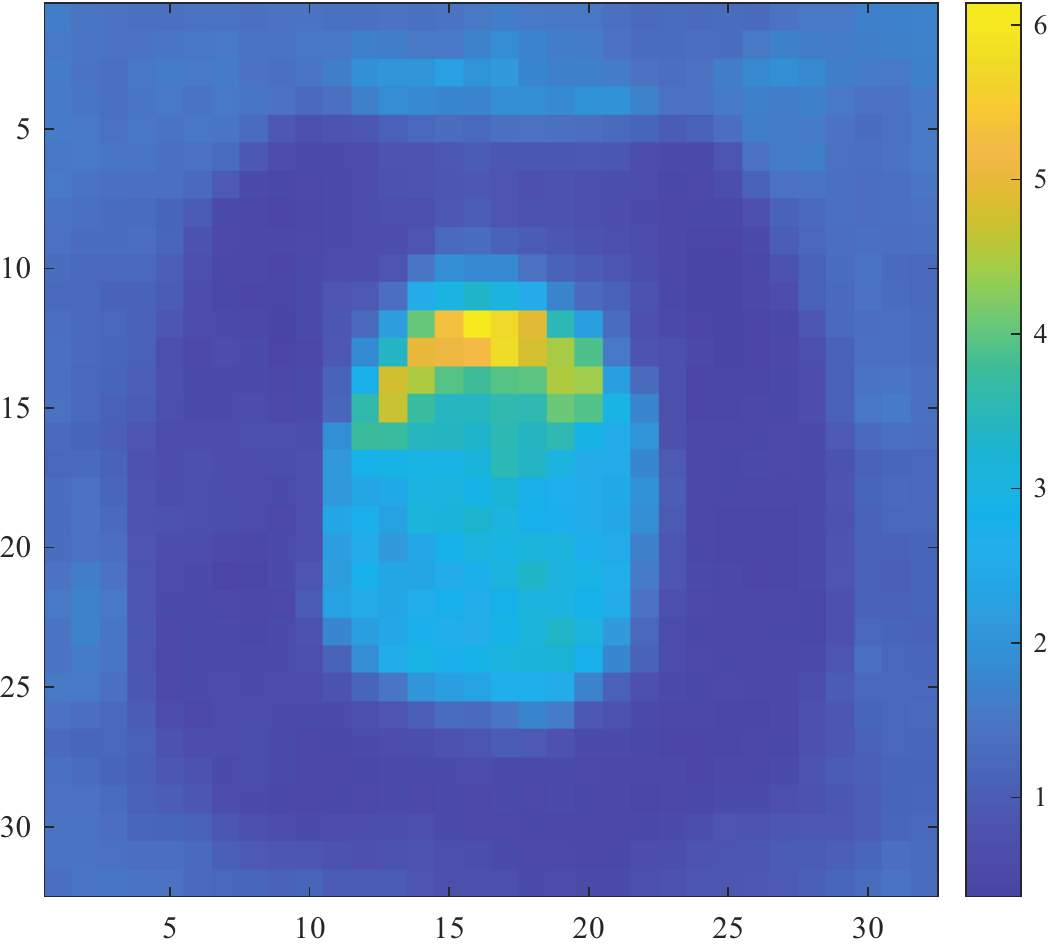}\\
\end{center}
\end{minipage}\\ \\

\begin{minipage}{5mm}
\begin{center}
$\bm{\Phi}_C$
\end{center}
\end{minipage}&

\begin{minipage}{40mm}
\begin{center}
\includegraphics[width=40mm]{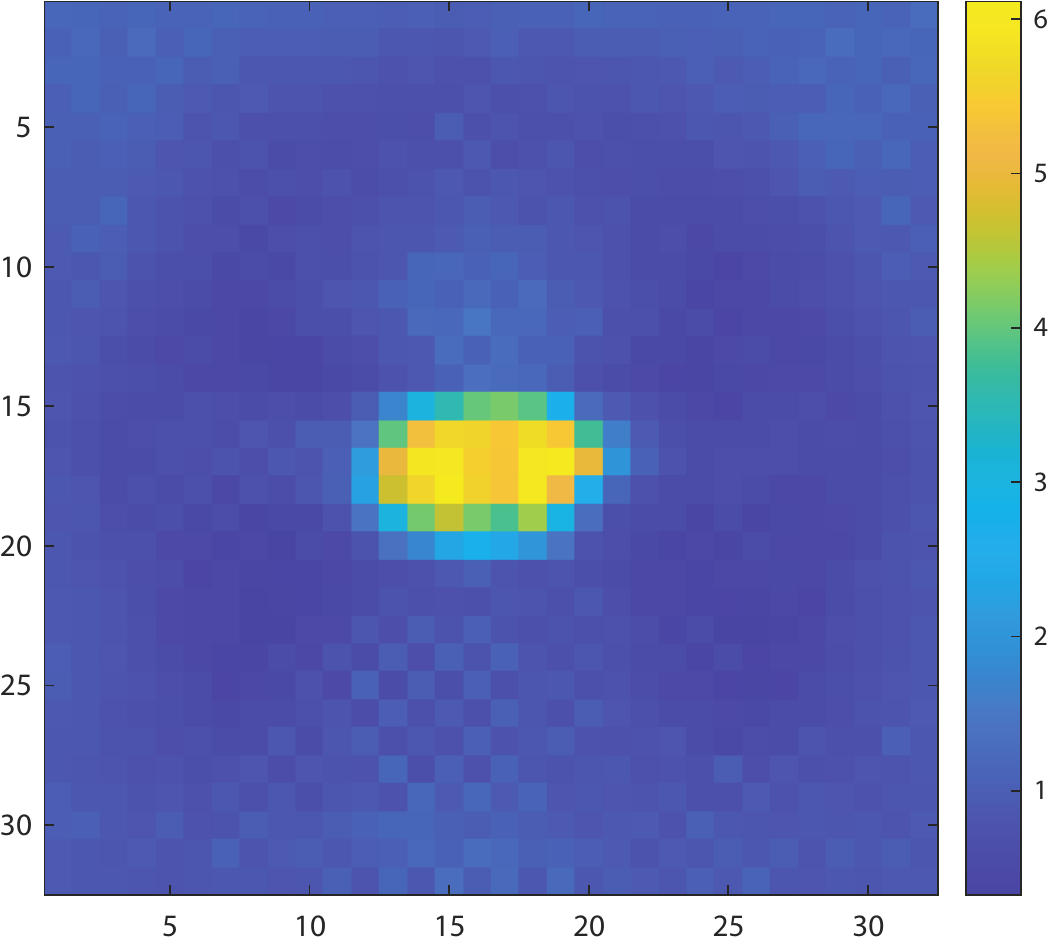}\\
\end{center}
\end{minipage}&

\begin{minipage}{40mm}
\begin{center}
\includegraphics[width=40mm]{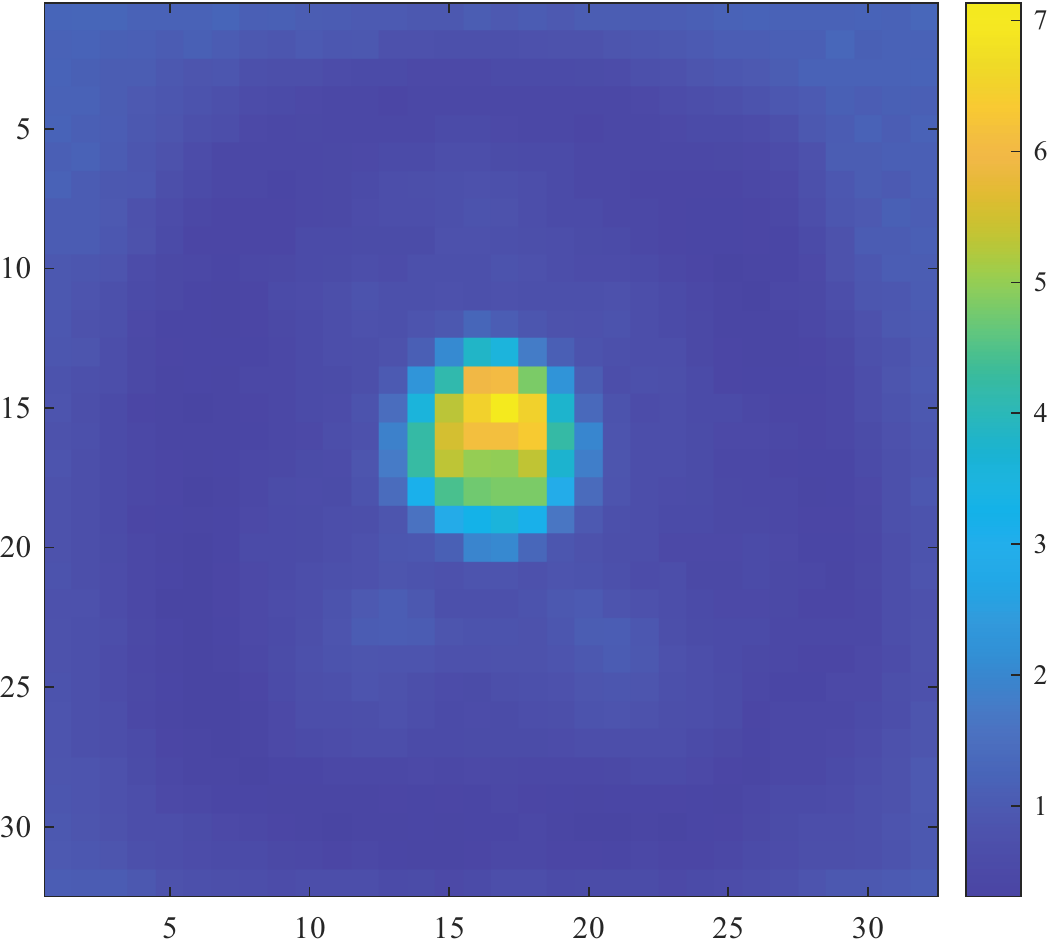}\\
\end{center}
\end{minipage}&

\begin{minipage}{40mm}
\begin{center}
\includegraphics[width=40mm]{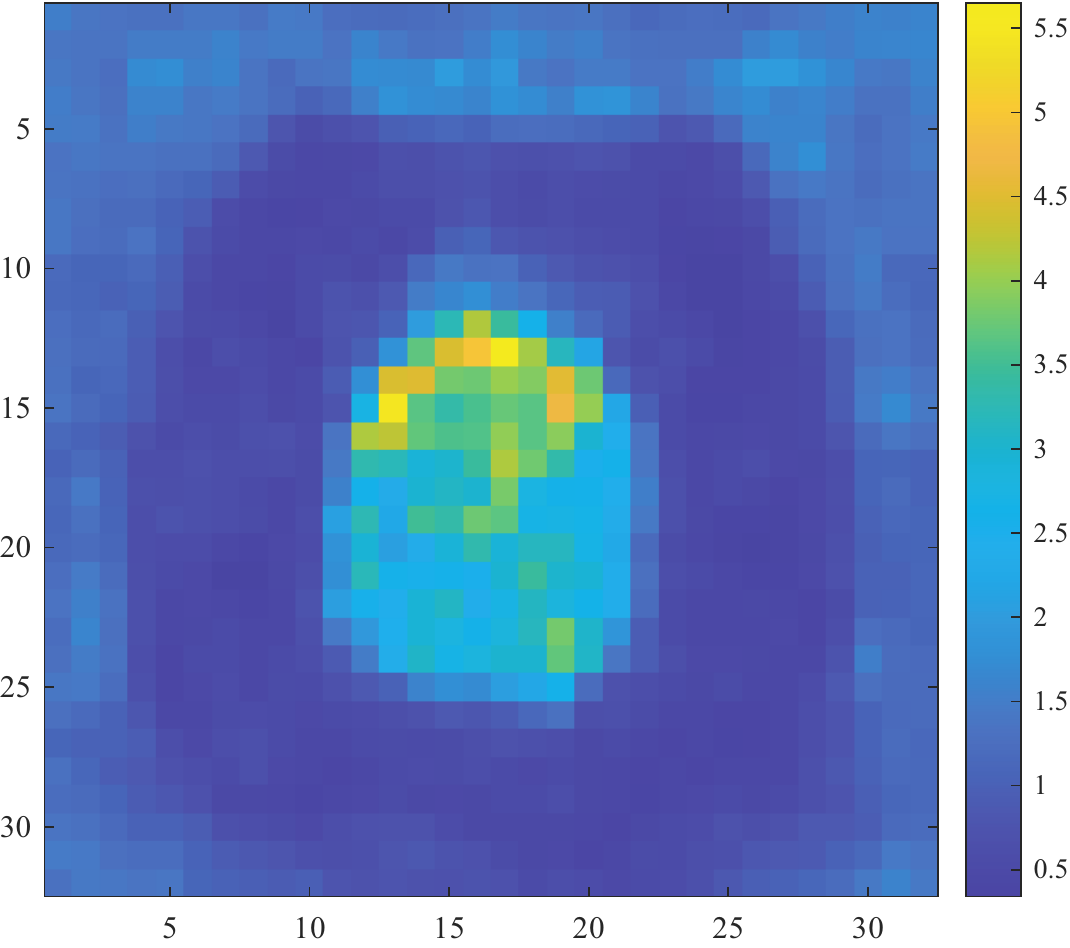}\\
\end{center}
\end{minipage}

\end{tabular}
\end{center}
\caption{Distance-accumulation images $\bm{\Phi}_M$, $\bm{\Phi}_{KL}$, $\bm{\Phi}_B$, and $\bm{\Phi}_C$ for the cases of airplanes and dogs, birds and dogs, cats and dogs, respectively.}
\label{fig:acc_image}
\end{figure}

As shown in Figure \ref{fig:dis_matrix},  $\bm{D}_M$ and $\bm{D}_{KL}$ for all of three cases appear chaotic and uninformative. By contrast, the local distances with high values represent a grid-like pattern in the middle of $\bm{D}_B$ and $\bm{D}_C$, where the exponent coefficient $s$ in $\bm{D}_C$ was set to $0.3$. These high-valued local distances can effectively be used to distinguish the corresponding elements of the random vectors $X_1$ and $X_2$ in the measurable space, which are also regarded as the corresponding pixels of the images $\bm{a}_k$ and $\bm{b}_k$. Considering that $\bm{D}_B$ is a particular case of $\bm{D}_C$ where $s$ is set to $1/2$, the $\bm{D}_C$-like statistical distance matrices are confirmed to be valid.

Furthermore, for each image pixel, we can accumulate all of its related local distances as a value and assign it to the current pixel to form a distance-accumulation vector $\bm{\phi} := [\acute{\phi}_{t}]\ \in \mathbb{R}^{d^2 \times 1},\ t = 1,\ldots,d^2$ by using the following distance-matrix-imaging method:
\begin{align}
  \acute{\phi}_{t}
  =& \sum_{u=1}^{d^2}\delta_{ut} + \sum_{v=1}^{d^2}\delta_{tv}.
\end{align}
Then, $\bm{\phi}$ is re-formed as a distance-accumulation image $\bm{\Phi} := [\phi_{ij}] = \widehat{\ve}\bm{\phi},\ i,j = 1,\ldots, d$, where $\widehat{\ve}(\cdot)$ denotes the reverse process of $\ve(\cdot)$. The effects of the statistical distance matrices are more clearly reflected in their distance-accumulation images than themselves. As illustrated in Figure \ref{fig:acc_image}, the distance-accumulation images of $\bm{D}_M$ and $\bm{D}_D$ are disordered, and those of $\bm{D}_B$ and $\bm{D}_C$ are ordered. The main representation is that the high-value pixels are all concentrated in the center of the distance-accumulation image, showing a distribution similar to a circle or an ellipse. For objects with low similarities, such as airplanes and dogs, the number of high-value pixels is greater and their locations are more concentrated. By contrast, between objects with high similarities, such as cats and dogs, high-value pixels are fewer and concentrated in the center of the image more broadly along with middle-value pixels. Therefore, the distance-matrix-imaging method can simultaneously quantify the differences between the pixels of every two objects in degree and position.

\begin{figure}[H]
\begin{center}
\begin{tabular}{c}

\begin{minipage}{77mm}
\begin{center}
\includegraphics[width=77mm]{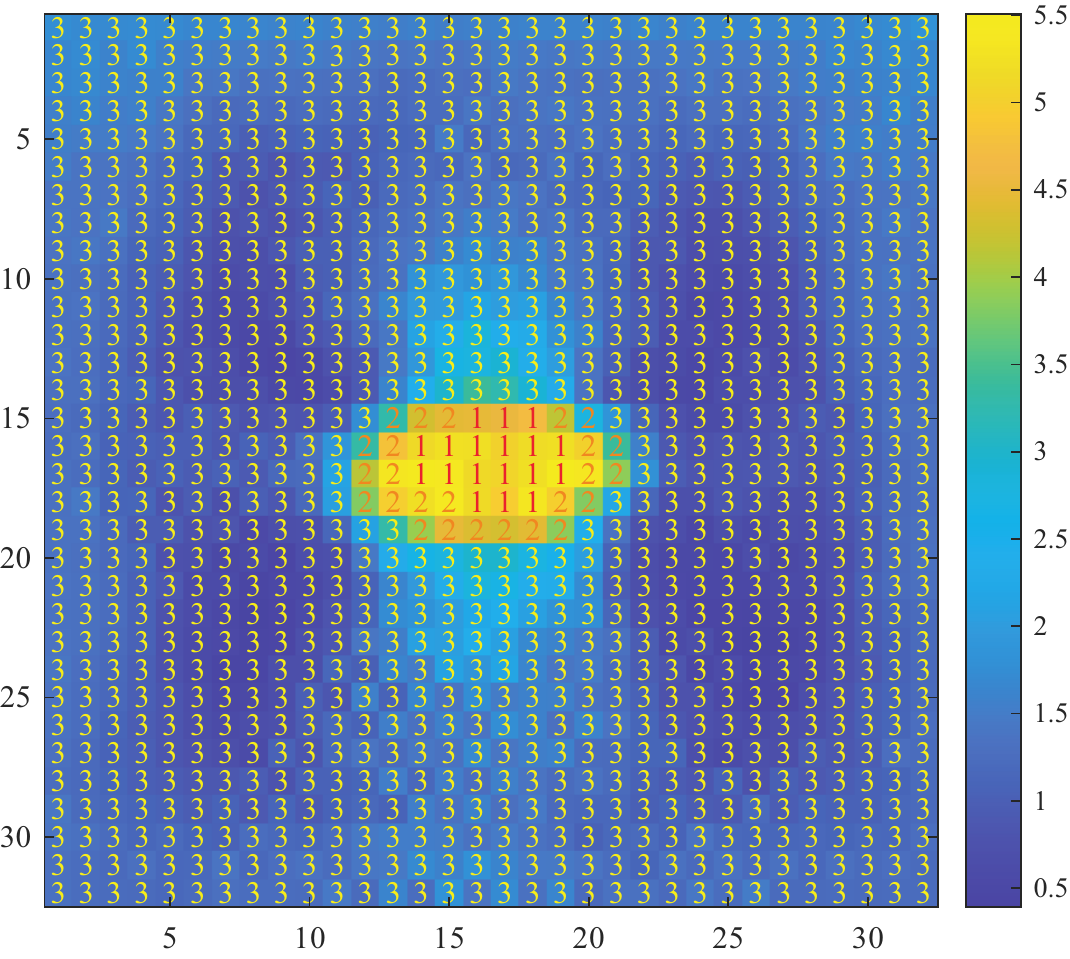}\\
(a)
\end{center}
\end{minipage}

\begin{minipage}{77mm}
\begin{center}
\includegraphics[width=77mm]{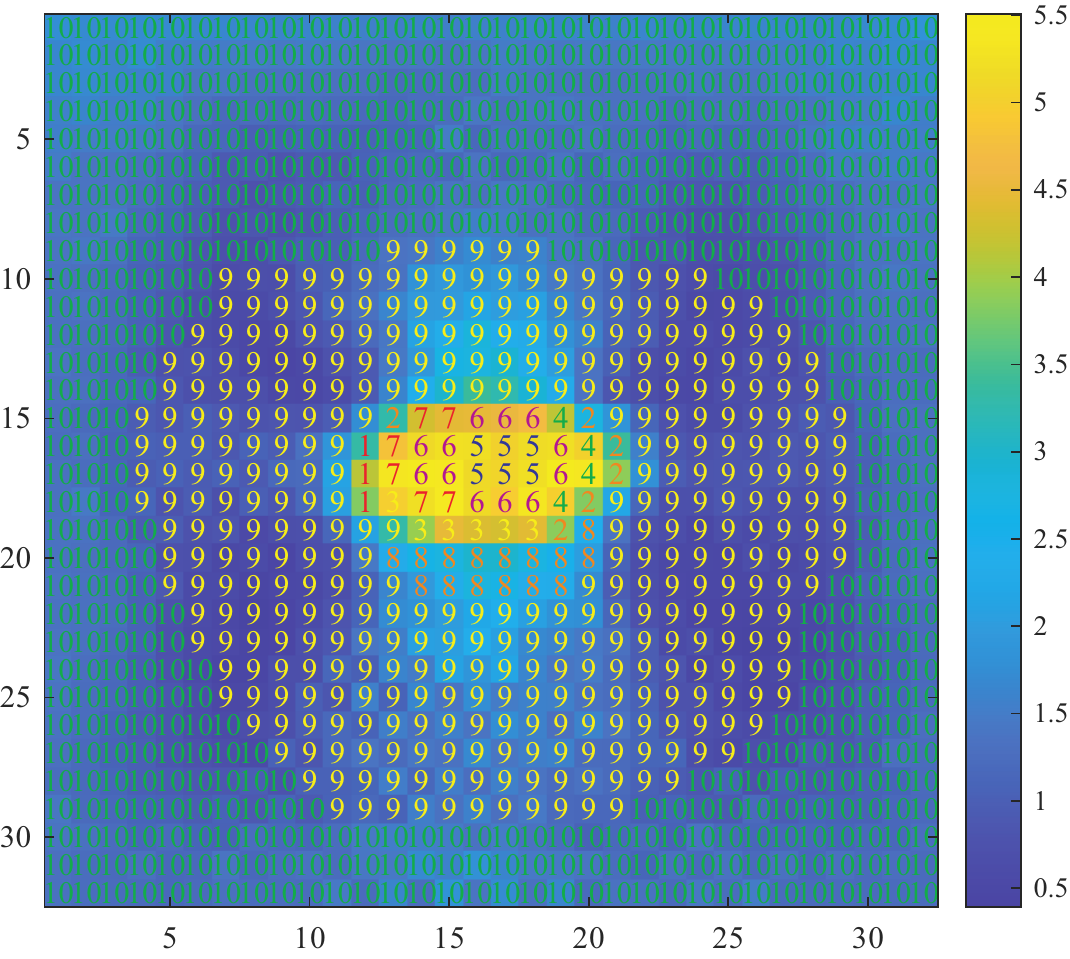}\\
(b)
\end{center}
\end{minipage}

\end{tabular}
\end{center}
\caption{The labeled distance-accumulation image $\bm{\Phi}_B$ to show the hierarchical clustering results for the case of airplanes and dogs using the statistical distance matrix $\bm{D}_B$. (a): three clusters; (b): ten clusters.}
\label{fig:hie_cluster_ad}
\end{figure}

\begin{figure}[h]
\begin{center}
\begin{tabular}{c}

\begin{minipage}{77mm}
\begin{center}
\includegraphics[width=77mm]{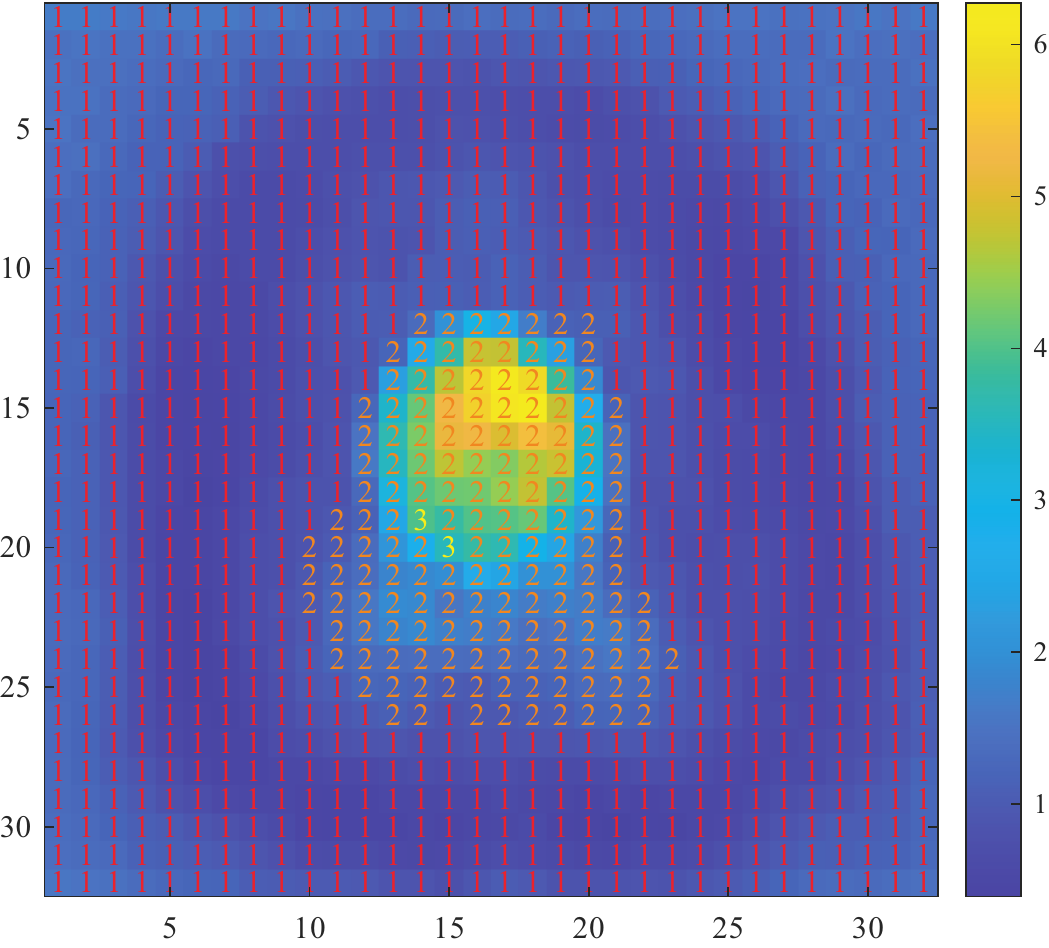}\\
(a)
\end{center}
\end{minipage}

\begin{minipage}{77mm}
\begin{center}
\includegraphics[width=77mm]{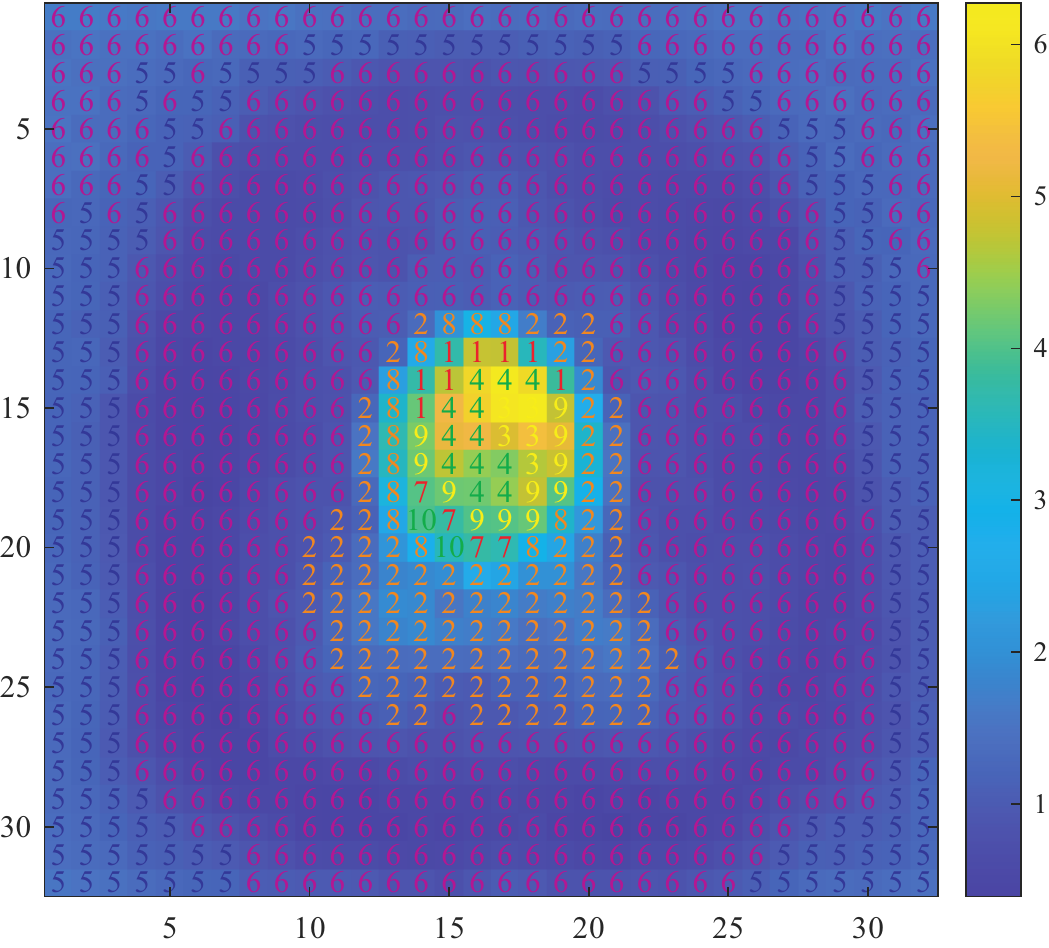}\\
(b)
\end{center}
\end{minipage}

\end{tabular}
\end{center}
\caption{The labeled distance-accumulation image $\bm{\Phi}_B$ to show the hierarchical clustering results for the case of birds and dogs using the statistical distance matrix $\bm{D}_B$. (a): three clusters; (b): ten clusters.}
\label{fig:hie_cluster_bd}
\end{figure}

\begin{figure}[h]
\begin{center}
\begin{tabular}{c}

\begin{minipage}{77mm}
\begin{center}
\includegraphics[width=77mm]{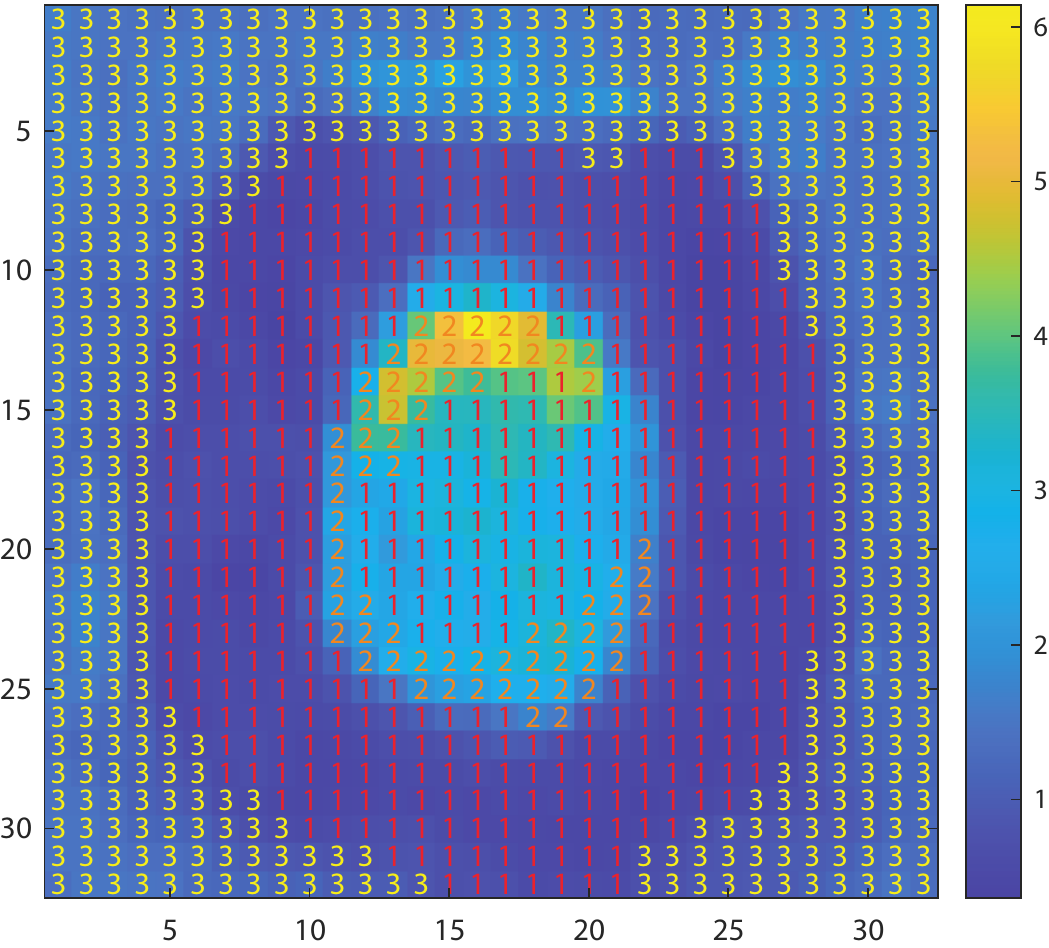}\\
(a)
\end{center}
\end{minipage}

\begin{minipage}{77mm}
\begin{center}
\includegraphics[width=77mm]{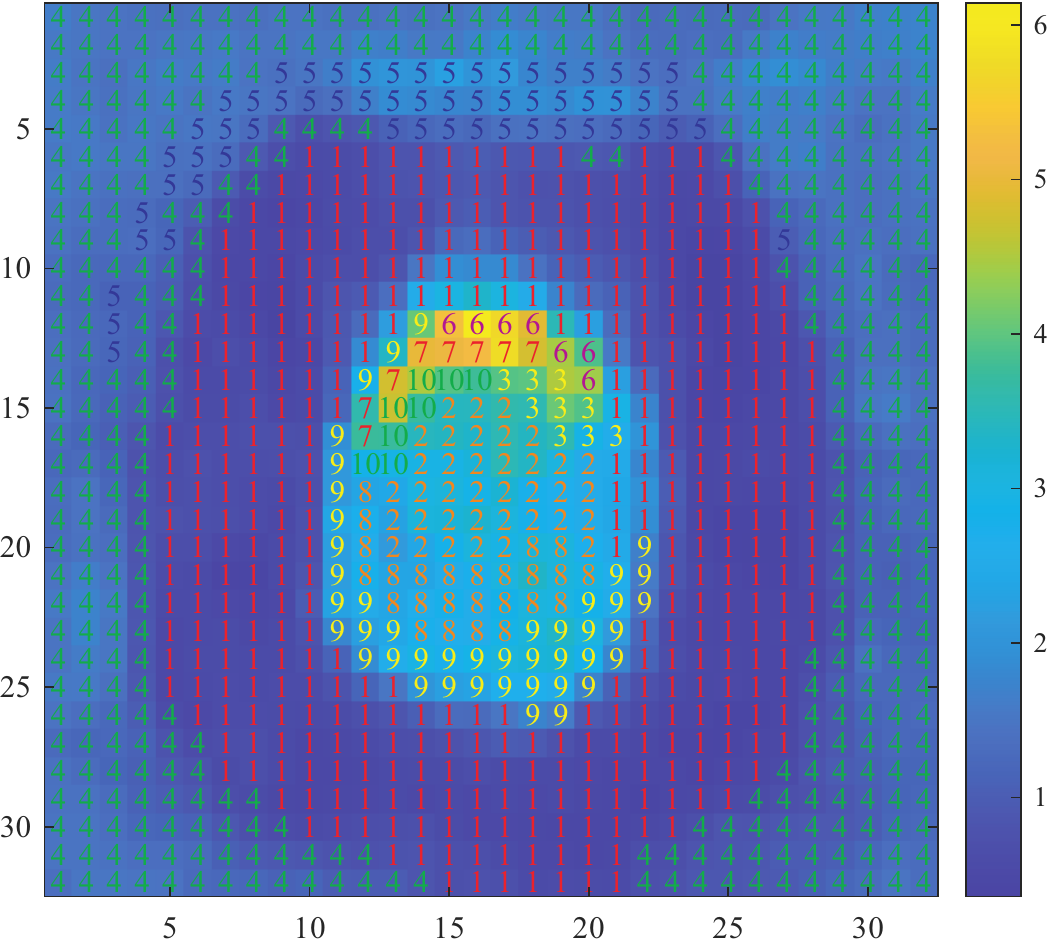}\\
(b)
\end{center}
\end{minipage}

\end{tabular}
\end{center}
\caption{The labeled distance-accumulation image $\bm{\Phi}_B$ to show the hierarchical clustering results for the case of cats and dogs using the statistical distance matrix $\bm{D}_B$. (a): three clusters; (b): ten clusters.}
\label{fig:hie_cluster_cd}
\end{figure}

Finally, the hierarchical clustering algorithm mentioned above is used to cluster the pixels of object images based on the statistical distance matrix $\bm{D}_B$. In Figures \ref{fig:hie_cluster_ad}, \ref{fig:hie_cluster_bd}, and \ref{fig:hie_cluster_cd}, the pixels of object images are separated into three and ten clusters and labeled in the distance-accumulation images for the three cases. The cluster patterns are all circular or square, symmetrical, and radiate from a center point. Therefore, the cluster patterns can be considered as Mandalas \footnote{The word Mandala is a Sanskrit term meaning “sacred circle.” In various religious traditions, such as Hinduism, Buddhism, Jainism, and Shintoism, a mandala is used as a map to represent paradise, gods, or actual shrines. Mandalas are circular or square and designed with repeating colors, shapes, and patterns that radiate from a center point. Mandalas can be precise, carefully measured, geometric, and perfectly symmetrical.}. Thus, we establish the term “Information Mandala” to describe the statistical distance matrix with clustering.

\section{Discussions}

We first discuss why the statistical distance matrices $\bm{D}_B$ and $\bm{D}_C$ are effective in the feature-distance measurement. There is a quadratic term with respect to the mean vectors $\bm{\mu}_1$ and $\bm{\mu}_2$ and covariance matrices  $\bm{\Sigma}_1$ and $\bm{\Sigma}_2$ in $\bm{D}_B$ and $\bm{D}_C$. This term also exists in $\bm{D}_M$ and $\bm{D}_{KL}$. There is also an item that contains only the logarithm of the covariance ratio in $\bm{D}_B$ and $\bm{D}_C$ and does not contain the mean vector. There is no such item containing only the covariance matrix in $\bm{D}_M$, and the only item containing the covariance ratio in $\bm{D}_{KL}$ has not been calculated logarithmically. When the two mean vectors are equal or approximate, the value of the quadratic term tends to zero, meaning that when the two probability distributions overlap heavily, the term of the covariance ratio plays a more important role than the quadratic term. This is the reason $\bm{D}_B$ and $\bm{D}_C$ are effective.

Next, we explain why cluster analysis is needed. Before cluster analysis, the statistical distance matrix $\bm{D}$ represents all the local distances between the elements of the random vector $X$. A large number of the local distances are so near zero that $\bm{D}$ becomes sparse. On the other hand, in graph theory, $\bm{D}$  can be mapped to a directed graph, whose vertices are defined as the elements of $X$, and whose edges are assigned to the local distances. However, in some applications when this graph is large, processing the edges with small values increases the computational complexity. Therefore, a tree structure is applied to reduce the number of unimportant edges in the graph and to arrange the vertices hierarchically according to their important edges with reassigned values. This tree structure can greatly accelerate the access speed and save memory space in the computer.

Finally, we consider the relation between the statistical distance matrices and the Capsule Neural Network (CapsNet), which is a novel and useful model of neural networks proposed in \cite{Sabour17, Hinton18}. In this paper, the statistical distance matrices are similar to the matrix of weights in CapsNet. Thus, they are viewpoint-invariant and can be used to distinguish an object no matter how much its pose has changed in the image. However, compared to CapsNet, the proposed statistical distance matrices are more intuitive based on the distance-accumulation images and more specifiable by using the hierarchical clustering method. Therefore, the statistical distance matrices, represented as Information Mandalas, can be considered an extended version of the matrix of weights.

\section{Conclusion}

Through the experimental comparisons of object images whose pixels are considered features, we confirmed that $\bm{D}_C$-like statistical distance matrices are more effective in distinguishing objects than other distance matrices. Their distance-accumulation images showed that high-valued pixels were concentrated in the middle of the image. Moreover, we found that after the hierarchical clustering of the distance matrix, all the pixel clusters basically surround the center of the image and are arranged radially from inside to outside according to the distance value. Since these patterns are very similar to Mandalas, we refer to the statistical distance matrix with clustering as the Information Mandala. The Information Mandala is a new form of entropy, which is an important means to understand convolutional neural networks.


\appendix

\section{Derivation for the Bhattacharyya distance $D_B$ (Eqn.\ (\ref{normal_b}))}\label{sec:appendix}

Let two $d$-dimensional random vectors $X_1$ and $X_2$ follow two normal distributions $\mathcal{N}(\bm{\mu}_1, \bm{\Sigma}_1)$ and $\mathcal{N}(\bm{\mu}_2, \bm{\Sigma}_2)$, respectively.
Their corresponding probability density functions $p(\bm{x})$ and $q(\bm{x})$ are defined as
\begin{align}
    p(\bm{x})
    :=& \det(2\pi\bm{\Sigma}_1)^{-\frac{1}{2}} \exp\left[ -\frac{1}{2}(\bm{x}-\bm{\mu}_1)^\intercal\bm{\Sigma}_1^{-1}(\bm{x}-\bm{\mu}_1) \right], \nonumber \\
    q(\bm{x})
    :=& \det(2\pi\bm{\Sigma}_2)^{-\frac{1}{2}} \exp\left[ -\frac{1}{2}(\bm{x}-\bm{\mu}_2)^\intercal\bm{\Sigma}_2^{-1}(\bm{x}-\bm{\mu}_2) \right].
\end{align}
The product of their square roots is written as
\begin{align}
    p^{\frac{1}{2}}(\bm{x})q^{\frac{1}{2}}(\bm{x})
    =& \det(2\pi\bm{\Sigma}_1)^{-\frac{1}{4}}\det(2\pi\bm{\Sigma}_2)^{-\frac{1}{4}} \nonumber \\
    &  \times \exp\left[ -\frac{1}{4}(\bm{x}-\bm{\mu}_1)^\intercal\bm{\Sigma}_1^{-1}(\bm{x}-\bm{\mu}_1) -\frac{1}{4}(\bm{x}-\bm{\mu}_2)^\intercal\bm{\Sigma}_2^{-1}(\bm{x}-\bm{\mu}_2)\right].  \label{eqn_0}
\end{align}
By integrating Eqn. (\ref{eqn_0}) in $\mathbb{R}^{d}$ with respect to $\bm{x}$, we obtain
\begin{align}
    \int_{\mathbb{R}^{d}} p_{X}^{\frac{1}{2}}(\bm{x})q_{X}^{\frac{1}{2}}(\bm{x}) d\bm{x}
    =& \det(2\pi\bm{\Sigma}_1)^{-\frac{1}{4}}\det(2\pi\bm{\Sigma}_2)^{-\frac{1}{4}} \label{eqn_1}\\
    &  \times \exp \left[ -\frac{1}{4}(\bm{\mu}_1^\intercal\bm{\Sigma}_1^{-1}\bm{\mu}_1 + \bm{\mu}_2^\intercal\bm{\Sigma}_2^{-1}\bm{\mu}_2)\right] \label{eqn_2}\\
    &  \times \int_{\mathbb{R}^{d}} \exp\left[ -\frac{1}{4}\bm{x}^\intercal(\bm{\Sigma}_1^{-1} + \bm{\Sigma}_2^{-1})\bm{x}
        +\frac{1}{2}(\bm{\mu}_1^\intercal\bm{\Sigma}_1^{-1} + \bm{\mu}_2^\intercal\bm{\Sigma}_2^{-1})\bm{x} \right]  d\bm{x}. \label{eqn_3}
\end{align}
Eqns.\ (\ref{eqn_1}), (\ref{eqn_2}), and (\ref{eqn_3}) can be transformed as follows. First, we define the following term:
\begin{align}
     &  \exp \left[ \frac{1}{8}(\bm{\mu}_1^\intercal\bm{\Sigma}_1^{-1} + \bm{\mu}_2^\intercal\bm{\Sigma}_2^{-1})(\bm{\Sigma}_1^{-1}\bm{\Sigma}\bm{\Sigma}_2^{-1})^{-1}(\bm{\Sigma}_1^{-1} \bm{\mu}_1+ \bm{\Sigma}_2^{-1}\bm{\mu}_2)\right] \nonumber \\
     =& \exp \left[ \frac{1}{8}(\bm{\mu}_1^\intercal\bm{\Sigma}_1^{-1}\bm{\Sigma}_2\bm{\Sigma}^{-1}\bm{\mu}_1 + 2\bm{\mu}_1^\intercal\bm{\Sigma}^{-1}\bm{\mu}_2 + \bm{\mu}_2^\intercal\bm{\Sigma}_2^{-1}\bm{\Sigma}_1\bm{\Sigma}^{-1}\bm{\mu}_2)\right] \nonumber \\
     =& \exp \left[ \frac{1}{8}(2\bm{\mu}_1^\intercal\bm{\Sigma}_1^{-1}(\bm{I} - \bm{\Sigma}_1(\bm{\Sigma}_1 + \bm{\Sigma}_2)^{-1})\bm{\mu}_1 + 2\bm{\mu}_1^\intercal\bm{\Sigma}^{-1}\bm{\mu}_2 + 2\bm{\mu}_2^\intercal\bm{\Sigma}_2^{-1}(\bm{I} - \bm{\Sigma}_2(\bm{\Sigma}_1 + \bm{\Sigma}_2)^{-1})\bm{\mu}_2)\right] \nonumber \\
     =& \exp \left[ \frac{1}{8}(2\bm{\mu}_1^\intercal\bm{\Sigma}_1^{-1}\bm{\mu}_1 - \bm{\mu}_1^\intercal\bm{\Sigma}^{-1}\bm{\mu}_1 + 2\bm{\mu}_1^\intercal\bm{\Sigma}^{-1}\bm{\mu}_2 + 2\bm{\mu}_2^\intercal\bm{\Sigma}_2^{-1}\bm{\mu}_2 - \bm{\mu}_2^\intercal\bm{\Sigma}^{-1}\bm{\mu}_2)\right]. \label{eqn_4}
\end{align}
This transformation is based on
\begin{align}
    (\bm{\Sigma}_1 + \bm{\Sigma}_2)^{-1}
    =& \bm{\Sigma}_1^{-1} - \bm{\Sigma}_1^{-1}(\bm{\Sigma}_2^{-1} + \bm{\Sigma}_1^{-1})^{-1}\bm{\Sigma}_1^{-1} \nonumber \\
    =& \bm{\Sigma}_2^{-1} - \bm{\Sigma}_2^{-1}(\bm{\Sigma}_1^{-1} + \bm{\Sigma}_2^{-1})^{-1}\bm{\Sigma}_2^{-1}, \label{eqn_42}
\end{align}
with
\begin{align}
    (\bm{\Sigma}_1^{-1} + \bm{\Sigma}_2^{-1})
    =   \bm{\Sigma}_1^{-1}\bm{\Sigma}_2\bm{\Sigma}_2^{-1} + \bm{\Sigma}_2^{-1}\bm{\Sigma}_1\bm{\Sigma}_1^{-1}
    =& \bm{\Sigma}_1^{-1}(\bm{\Sigma}_1 + \bm{\Sigma}_2)\bm{\Sigma}_2^{-1} \nonumber \\
    =& \bm{\Sigma}_2^{-1}(\bm{\Sigma}_1 + \bm{\Sigma}_2)\bm{\Sigma}_1^{-1}.
\end{align}
Here, Eqn. (\ref{eqn_42}) holds by 
\begin{align}
    (\bm{A} + \bm{C}\bm{B}\bm{C}^\intercal)^{-1}
    =& \bm{A}^{-1} - \bm{A}^{-1}\bm{C}(\bm{B}^{-1}+\bm{C}^\intercal\bm{A}^{-1}\bm{C})^{-1}\bm{C}^\intercal\bm{A}^{-1},
\end{align}
where $\bm{A}$, $\bm{B}$, $\bm{C}$ are all positive-definite matrices.
$\bm{\Sigma}$ is a mean of $\bm{\Sigma}_1$ and $\bm{\Sigma}_2$ as
\begin{align}
    \bm{\Sigma}
    :=& \frac{\bm{\Sigma}_1 + \bm{\Sigma}_2}{2}.
\end{align}
Then let Eqn. (\ref{eqn_3}) be
\begin{align}
       & \int_{\mathbb{R}^{d}} \exp\left[ -\frac{1}{2}\bm{x}^\intercal\bm{\Sigma}^{-1}\bm{x}
         +\frac{1}{2}(\bm{\mu}_1^\intercal\bm{\Sigma}_1^{-1} + \bm{\mu}_2^\intercal\bm{\Sigma}_2^{-1})\bm{x} \right]  d\bm{x} \nonumber \\
     =& \int_{\mathbb{R}^{d}} \exp\left[ -\frac{1}{2}\bm{y}^\intercal\bm{y}
         +\frac{1}{2}(\bm{\mu}_1^\intercal\bm{\Sigma}_1^{-1} + \bm{\mu}_2^\intercal\bm{\Sigma}_2^{-1})(\bm{\Sigma}_1^{-1}\bm{\Sigma}\bm{\Sigma}_2^{-1})^{-\frac{1}{2}}\bm{y} \right]  d\left[(\bm{\Sigma}_1^{-1}\bm{\Sigma}\bm{\Sigma}_2^{-1})^{-\frac{1}{2}}\bm{y}\right], \nonumber \\
      =& \underbrace{\mathrm{Eqn}.\ (\ref{eqn_4})}_{\text{first factor}} \times \underbrace{\mathrm{Eqn}.\ (\ref{eqn_3})   / \mathrm{Eqn}.\ (\ref{eqn_4})}_{\text{second factor}}, \label{eqn_41}
\end{align}
where 
\begin{align}
     \bm{y} 
     :=& (\bm{\Sigma}_1^{-1}\bm{\Sigma}\bm{\Sigma}_2^{-1})^{\frac{1}{2}}\bm{x}. \label{eqn_43}
\end{align}
By multiplying Eqn. (\ref{eqn_2}) and the first factor of Eqn. (\ref{eqn_41}) together, we obtain
\begin{align}
    \mathrm{Eqn}.\ (\ref{eqn_2}) \times \mathrm{Eqn}.\ (\ref{eqn_4})
    =& \exp \left[ -\frac{1}{8}(\bm{\mu}_1^\intercal\bm{\Sigma}^{-1}\bm{\mu}_1 - 2\bm{\mu}_1^\intercal\bm{\Sigma}^{-1}\bm{\mu}_2 + \bm{\mu}_2^\intercal\bm{\Sigma}^{-1}\bm{\mu}_2)\right] \nonumber \\
    =& \exp \left[-\frac{1}{8}(\bm{\mu}_1 - \bm{\mu}_2)^\intercal \bm{\Sigma}^{-1} (\bm{\mu}_1 - \bm{\mu}_2))\right]. \label{eqn_5}
\end{align}
The second factor of Eqn. (\ref{eqn_41}) is transformed as
\begin{align}
    & \det \left[ (\bm{\Sigma}_1^{-1}\bm{\Sigma}\bm{\Sigma_2}^{-1})^{-\frac{1}{2}} \right]
    \int_{\mathbb{R}^{d}} \exp \left\{ -\frac{1}{2} \left[\bm{y}-\frac{1}{2}(\bm{\Sigma}_1^{-1}\bm{\Sigma}\bm{\Sigma_2}^{-1})^{-\frac{1}{2}}(\bm{\Sigma}_1^{-1}\bm{\mu}_1+\bm{\Sigma}_2^{-1}\bm{\mu}_2)\right]^\intercal \right. \nonumber \\
    & \times \left.\left[\bm{y}-\frac{1}{2}(\bm{\Sigma}_1^{-1}\bm{\Sigma}\bm{\Sigma_2}^{-1})^{-\frac{1}{2}}(\bm{\Sigma}_1^{-1}\bm{\mu}_1+\bm{\Sigma}_2^{-1}\bm{\mu}_2)\right] \right\} d\bm{y} \nonumber \\
    =& (2\pi)^{\frac{k}{2}} \det (\bm{\Sigma}_1^{-1}\bm{\Sigma}\bm{\Sigma_2}^{-1})^{-\frac{1}{2}} \label{eqn_6}
\end{align}
using change-of-variable technique. Given
\begin{align}
    \mathrm{Eqn}.\ (\ref{eqn_1}) \times \mathrm{Eqn}.\ (\ref{eqn_6})
    =& \det(2\pi\bm{\Sigma}_1)^{-\frac{1}{4}}\det(2\pi\bm{\Sigma}_2)^{-\frac{1}{4}}
        (2\pi)^{\frac{k}{2}}\det (\bm{\Sigma}_1^{-1}\bm{\Sigma}\bm{\Sigma_2}^{-1})^{-\frac{1}{2}}  \nonumber \\
    =& (2\pi)^{-\frac{k}{4}}\det \bm{\Sigma}_1^{-\frac{1}{4}} (2\pi)^{-\frac{k}{4}}\det \bm{\Sigma}_2^{-\frac{1}{4}}
        (2\pi)^{\frac{k}{2}} \det(\bm{\Sigma}_1^{-1}\bm{\Sigma}\bm{\Sigma_2}^{-1})^{-\frac{1}{2}}  \nonumber \\
    =& \det\bm{\Sigma}_1^{\frac{1}{4}} \det\bm{\Sigma}^{-\frac{1}{2}} \det\bm{\Sigma}_2^{\frac{1}{4}}, \label{eqn_7}
\end{align}
the Bhattacharyya distance $D_B$ is achieved by
\begin{align}
    -\ln (\mathrm{Eqn}.\ (\ref{eqn_5}) \times \mathrm{Eqn}.\ (\ref{eqn_7}))
    = \frac{1}{8}(\bm{\mu}_1 - \bm{\mu}_2)^\intercal \bm{\Sigma}^{-1} (\bm{\mu}_1 - \bm{\mu}_2)
      + \frac{1}{2} \ln\left[\det\bm{\Sigma}_1^{-\frac{1}{2}} \det\bm{\Sigma} \det\bm{\Sigma}_2^{-\frac{1}{2}}\right].
\end{align}

\bibliographystyle{plain}
\bibliography{ref}

\end{document}